\documentclass{article}

\PassOptionsToPackage{numbers, compress}{natbib}

\usepackage[preprint]{neurips_2024}

\usepackage[utf8]{inputenc} 
\usepackage[T1]{fontenc}    
\usepackage{hyperref}       
\usepackage{url}            
\usepackage{booktabs}       
\usepackage{amsfonts}       
\usepackage{nicefrac}       
\usepackage{microtype}      
\usepackage{graphicx}
\usepackage{multirow} 
\usepackage{makecell} 
\usepackage{colortbl} 
\usepackage{diagbox} 
\usepackage{siunitx}
\usepackage{enumitem} 
\usepackage{arydshln} 
\usepackage{tabularx}
\usepackage{booktabs} 
\usepackage{xspace} 
\usepackage{pifont} 
\usepackage[dvipsnames]{xcolor} 
\usepackage{dsfont} 
\usepackage{amsmath}
\usepackage{cleveref}
\usepackage{bm}

\def\eg{\emph{e.g.}\xspace} 
\def\ie{\emph{i.e.}\xspace}

\crefname{figure}{fig.}{figs.}
\Crefname{figure}{Fig.}{Figs.}
\crefname{table}{tab.}{tables.}
\Crefname{table}{Tab.}{Tables.}

\newcommand{\ours}{Splat-SLAM\xspace}
\newcommand{\boldparagraph}[1]{\vspace{0pt}\noindent{\bf #1}}

\newcommand{\greencheck}{{\color{OliveGreen}\checkmark}}
\newcommand{\redx}{{\color{red}\ding{55}}}

\colorlet{colorFst}{Green!25}       
\colorlet{colorSnd}{SpringGreen!45} 
\colorlet{colorTrd}{Yellow!30}      
\colorlet{colorLow}{darkgray!30}    
\newcommand{\fst}{\cellcolor{colorFst}\bf}   
\newcommand{\nd}{\cellcolor{colorSnd}}      
\newcommand{\rd}{\cellcolor{colorTrd}}      

\definecolor{gray}{rgb}{0.65,0.65,0.65}
\definecolor{mycol}{rgb}{0.90,0.95,1.0}

\title{\ours: Globally Optimized RGB-only SLAM with 3D Gaussians}

\author{%
  Erik Sandström\thanks{This work was conducted during an internship at Google.} \\ Google \\ ETH Zürich \And Keisuke Tateno \\ Google \And Michael Oechsle \\ Google \And Michael Niemeyer \\ Google \And Luc Van Gool \\ ETH Zürich \\ INSAIT \And Martin R. Oswald \\ ETH Zürich \\ University of Amsterdam \And Federico Tombari \\ Google \\ TU München   
}

\begin{document}

\maketitle

\begin{abstract}
3D Gaussian Splatting has emerged as a powerful representation of geometry and appearance for RGB-only dense Simultaneous Localization and Mapping (SLAM), as it provides a compact dense map representation while enabling efficient and high-quality map rendering. 
However, existing methods show significantly worse reconstruction quality than competing methods using other 3D representations, \eg neural points clouds, since they either do not employ global map and pose optimization or make use of monocular depth.
In response, we propose the first RGB-only SLAM system with a dense 3D Gaussian map representation that utilizes all benefits of globally optimized tracking by adapting dynamically to keyframe pose and depth updates by actively deforming the 3D Gaussian map. Moreover, we find that refining the depth updates in inaccurate areas with a monocular depth estimator further improves the accuracy of the 3D reconstruction. Our experiments on the Replica, TUM-RGBD, and ScanNet datasets indicate the effectiveness of globally optimized 3D Gaussians, as the approach achieves superior or on par performance with existing RGB-only SLAM methods methods in tracking, mapping and rendering accuracy while yielding small map sizes and fast runtimes. The source code is available at \url{https://github.com/eriksandstroem/Splat-SLAM}.

\end{abstract}

\section{Introduction}

A common factor within the recent trend of dense SLAM is that the majority of works reconstruct a dense map by optimizing a neural implicit encoding of the scene, either as weights of an MLP~\cite{azinovic2022neural,Sucar2021IMAP:Real-Time,matsuki2023imode,ortiz2022isdf}, as features anchored in dense grids~\cite{zhu2022nice,newcombe2011kinectfusion,Weder2020RoutedFusion,weder2021neuralfusion,sun2021neuralrecon,bovzivc2021transformerfusion,li2022bnv,zou2022mononeuralfusion,uncleslam2023}, using hierarchical octrees~\cite{yang2022vox}, via voxel hashing~\cite{zhang2023go,zhang2023hi,chung2022orbeez,Rosinol2022NeRF-SLAM:Fields,matsuki2023newton}, point clouds \cite{hu2023cp,sandstrom2023point,liso2024loopyslam,zhang2024glorie} or axis-aligned feature planes~\cite{mahdi2022eslam,peng2020convolutional}. We have also seen the introduction of 3D Gaussian Splatting (3DGS) to the dense SLAM field~\cite{yugay2023gaussianslam,keetha2023splatam,yan2023gs,matsuki2023gaussian,huang2023photo}.

Out of this 3D representation race there is, however, not yet a clear winner. In the context of dense SLAM, a careful modeling choice needs to be made to achieve accurate surface reconstruction as well as low tracking errors. Some takeaways can be deduced from the literature: neural implicit point cloud representations achieve state-of-the-art reconstruction accuracy \cite{liso2024loopyslam,zhang2024glorie,sandstrom2023point}, especially with RGBD input. At the same time, 3D Gaussian splatting methods yield the highest fidelity renderings~\cite{matsuki2023gaussian, yugay2023gaussianslam, keetha2023splatam,huang2023photo,yan2023gs} and show promise in the RGB-only setting due to their flexibility in optimizing the surface location~\cite{huang2023photo,matsuki2023gaussian}. However, they are not leveraging any multi-view depth or geometric prior leading to poor geometry in the RGB-only setting. The majority of the aforementioned works \textit{only} deploy so called frame-to-model tracking, and do not implement global trajectory and map optimization, leading to excessive drift, especially in real world conditions. Instead, to this date, frame-to-frame tracking methods, coupled with loop closure and global bundle adjustment (BA) achieve state-of-the-art tracking accuracy~\cite{zhang2023go,zhang2023hi,zhang2024glorie}. However, they either use hierarchical feature grids~\cite{zhang2023go,zhang2023hi}, not suitable for map deformations at \eg loop closure as they require expensive reintegration strategies, or neural point clouds as in GlORIE-SLAM~\cite{zhang2024glorie}. While the neural point cloud is straightforward to deform, the depth guided rendering leads to artifacts when the depth is noisy and the surface estimation can only be adjusted locally since the point locations are not optimized directly.

\begin{figure}[t]
  \centering
  \footnotesize  
  \setlength{\tabcolsep}{2.5pt}
  \renewcommand{\arraystretch}{1}
  \newcommand{\sz}{0.24}
  \begin{tabular}{cccc} 
    \small Ground Truth &
    \small GlORIE-SLAM~\cite{zhang2024glorie} & 
    \small MonoGS~\cite{matsuki2023gaussian} & 
    \small \ours (Ours) \\
    \includegraphics[width=\sz\linewidth]{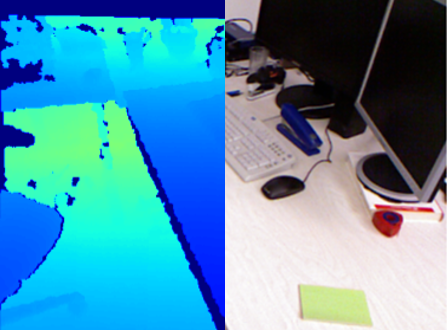} & 
    \includegraphics[width=\sz\linewidth]{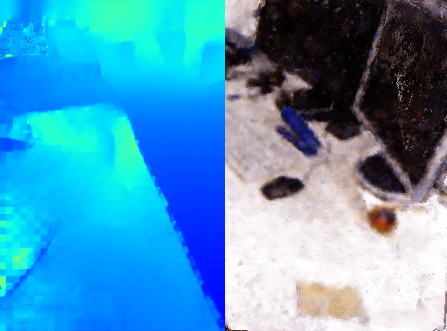} &
    \includegraphics[width=\sz\linewidth]{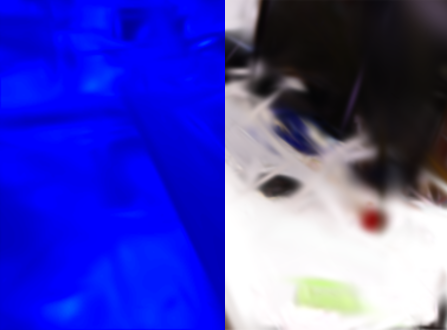} & 
    \includegraphics[width=\sz\linewidth]{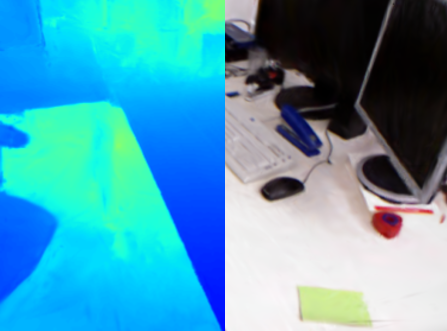} \\
    \begin{tabular}{@{}rl@{}} \hspace{-0.55cm} Depth L1 [cm]$\downarrow$ & PSNR$\uparrow$ \end{tabular} &
    \begin{tabular}{cc}  22.19 & \hspace{0.7cm} 18.78 \end{tabular} &
    \begin{tabular}{cc} 116.71 & \hspace{0.7cm} 18.41 \end{tabular} &
    \begin{tabular}{cc} \textbf{15.05} & \hspace{0.7cm} \textbf{24.06} \end{tabular} \\
    \midrule 
    \begin{tabular}{@{}rl@{}} ATE RMSE [cm]$\downarrow$\end{tabular} &
    \begin{tabular}{@{}c@{}} \textbf{4.2} \end{tabular} &
    \begin{tabular}{@{}c@{}} 76.56 \end{tabular} &
    \begin{tabular}{@{}c@{}} \textbf{4.2} \end{tabular} \\
    \midrule 
    \begin{tabular}{@{}rl@{}} Map Size [MB]$\downarrow$ \end{tabular} &
    \begin{tabular}{@{}c@{}} 382.4 \end{tabular} &
    \begin{tabular}{@{}c@{}} \textbf{5.2} \end{tabular} &
    \begin{tabular}{@{}c@{}} 10.8 \end{tabular}
  \end{tabular}
  \caption{\textbf{\ours.} Our system yields accurate scene reconstruction (rendering depth L1) and rendering (PSNR) and on par tracking accuracy (ATE RMSE) to GlORIE-SLAM and map size to MonoGS. The results averaged over all keyframes. The scene is from TUM-RGBD~\cite{Sturm2012ASystems} \texttt{fr1 room}.}
  \label{fig:teaser}
\end{figure}

In this work we propose an RGB-only SLAM system that combines the strengths of frame-to-frame tracking using recurrent dense optical flow~\cite{teed2021droid} with the fidelity of 3D Gaussians as the map representation~\cite{matsuki2023gaussian} (see \cref{fig:teaser}). The point-based 3D Gaussian map enables online map deformations at loop closure and global BA. To enable accurate surface reconstruction, we leverage consistent so called proxy depth that combines multi-view depth estimation with learned monocular depth. 
Our contribution comprises, for the first time, a SLAM pipeline encompassing all the following parts: 
\begin{itemize}[itemsep=0pt,topsep=2pt,leftmargin=10pt,label=$\bullet$]
\item A frame-to-frame RGB-only tracker with global consistency.
\item A dense deformable 3D Gaussian map that adapts online to loop closure and global BA.
\item A proxy depth map consisting of on-the-fly optimized multi-view depth and a monocular depth estimator leading to improved rendering and reconstruction quality.
\item Improved map sizes and runtimes compared to other dense SLAM approaches.
\end{itemize}


\section{Related Work} \label{sec:rel}

\boldparagraph{Dense Visual SLAM.} 
Curless and Levoy~\cite{curless1996volumetric} pioneered dense online 3D mapping with truncated signed distance functions, with KinectFusion~\cite{newcombe2011kinectfusion} demonstrating real-time SLAM via depth maps. 
Enhancements like voxel hashing~\cite{niessner2013voxel_hashing,Kahler2015infiniTAM,Oleynikova2017voxblox,dai2017bundlefusion,matsuki2023newton} and octrees~\cite{steinbrucker2013large,yang2022vox,marniok2017efficient,chen2013scalable,liu2020neural} improved scalability, while point-based SLAM~\cite{whelan2015elasticfusion,schops2019bad,cao2018real,Kahler2015infiniTAM,keller2013real,cho2021sp,zhang2020dense,sandstrom2023point,liso2024loopyslam,zhang2024glorie} has also been effective. To address pose drift, globally consistent pose estimation and dense mapping techniques have been developed, often dividing the global map into submaps~\cite{cao2018real,dai2017bundlefusion,fioraio2015large,tang2023mips,matsuki2023newton,maier2017efficient,kahler2016real,stuckler2014multi,choi2015robust,Kahler2015infiniTAM,reijgwart2019voxgraph,henry2013patch,bosse2003atlas,maier2014submap,tang2023mips,mao2023ngel,liso2024loopyslam}. Loop detection triggers submap deformation via pose graph optimization~\cite{cao2018real,maier2017efficient,tang2023mips,matsuki2023newton,kahler2016real,endres2012evaluation,engel2014lsd,kerl2013dense,choi2015robust,henry2012rgb,yan2017dense,schops2019bad,reijgwart2019voxgraph,henry2013patch,stuckler2014multi,wang2016online,matsuki2023newton,hu2023cp,mao2023ngel,liso2024loopyslam}. Sometimes global BA is used for refinement~\cite{dai2017bundlefusion,schops2019bad,cao2018real,teed2021droid,yan2017dense,yang2022fd,matsuki2023newton,chung2022orbeez,tang2023mips,hu2023cp}. 3D Gaussian SLAM with RGBD input has also been shown, but these methods do not consider global consistency via \eg loop closure~\cite{yugay2023gaussianslam,keetha2023splatam,yan2023gs}. Other approaches to global consistency minimize reprojection errors directly, with DROID-SLAM~\cite{teed2021droid} refining dense optical flow and camera poses iteratively, and recent enhancements like GO-SLAM~\cite{zhang2023go}, HI-SLAM~\cite{zhang2023hi}, and GlORIE-SLAM~\cite{zhang2024glorie} optimizing factor graphs for accurate tracking. For a recent survey on NeRF-inspired dense SLAM, see~\cite{tosi2024nerfs}.

\boldparagraph{RGB-only Dense Visual SLAM.}
The majority of NeRF inspired dense SLAM works using only RGB cameras do not address the problem of global map consistency or requires expensive reintegration strategies via backpropagation~\cite{Rosinol2022NeRF-SLAM:Fields, chung2022orbeez,li2023dense,zhu2023nicer,peng2024q,zhang2023go,zhang2023hi,hua2023fmapping,naumann2023nerf,hua2024hi}. Instead, the concurrent GlORIE-SLAM~\cite{zhang2024glorie} uses a feature based point cloud which can adapt to global map changes in a straight forward way. However, redundant points are not pruned, leading to large map sizes. Furthermore, the depth guided sampling during rendering leads to rendering artifacts when noise is present in the estimated depth. MonoGS~\cite{matsuki2023gaussian} and Photo-SLAM~\cite{huang2023photo} pioneered RGB-only SLAM with 3D Gaussians. However, they lack proxy depth which prevents them from achieving high accuracy mapping. MonoGS~\cite{matsuki2023gaussian} also lacks global consistency. Concurrent to our work, MoD-SLAM~\cite{zhou2024modslam} uses an MLP to parameterize the map via a unique reparameterization.




\begin{figure}[t!]
\centering
 \includegraphics[width=\linewidth]{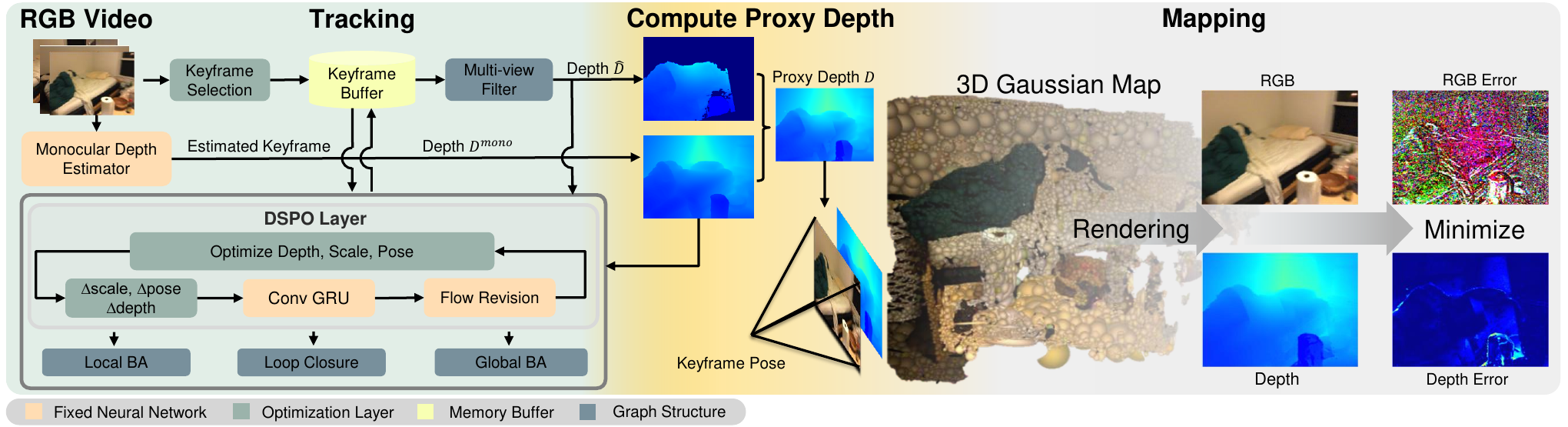}\\
\caption{\textbf{\ours{} Architecture.} 
Given an RGB input stream, we track and map each keyframe, initially estimating poses through local bundle adjustment (BA) using a DSPO (Disparity, Scale and Pose Optimization) layer. This layer integrates pose and depth estimation, enhancing depth with monocular depth. It further refines poses globally via online loop closure and global BA. The proxy depth map merges keyframe depths $\tilde{D}$ from the tracking with monocular depth $D^{mono}$ to fill gaps. Mapping employs a deformable 3D Gaussian map, optimizing its parameters through a re-rendering loss. Notably, the 3D map adjusts for global pose and depth updates before each mapping phase.
}
\label{fig:architecture}
\end{figure}

\section{Method} \label{sec:methods}
%
Splat-SLAM is a monocular SLAM system which tracks the camera pose while reconstructing the dense geometry of the scene in an online manner. This is achieved through the following steps: We first track the camera by performing local BA on selected keyframes by fitting them to dense optical flow estimates. The local BA optimizes the camera pose as well as the dense depth of the keyframe. For global consistency, when loop closure is detected, loop BA is performed on an extended graph including the loop nodes and edges (\cref{sec:tracking}). Interleaved with tracking, mapping is done on a progressively growing 3D Gaussian map which deforms online to the keyframe poses and so called proxy depth maps (\cref{sec:3dgs_rep}). For an overview of our method, see \cref{fig:architecture}.

\subsection{Tracking} \label{sec:tracking}
To predict the motion of the camera during scene exploration, we use a pretrained recurrent optical flow model~\cite{teed2020raft} coupled with a Disparity, Scale and Pose Optimization (DSPO) layer~\cite{zhang2024glorie} to jointly optimize camera poses and per pixel disparities.
In the following, we describe this process in detail.

Optimization is done with the Gauss-Newton algorithm over a factor graph $G(V,E)$, where the nodes $V$ store the keyframe pose and disparity, and edges $E$ store the optical flow between keyframes. Odometry keyframe edges are added to $G$ by computing the optical flow to the last added keyframe. If the mean flow is larger than a threshold $\tau\in\mathbb{R}$, the new keyframe is added to $G$. Edges for loop closure and global BA are discussed later. Importantly, the same objective is optimized for local BA, loop closure and global BA, but over factor graphs with different structures.

The DSPO layer consists of two optimization objectives that are optimized alternatively. The first objective, typically termed Dense Bundle Adjustment (DBA)~\cite{teed2021droid} optimizes the pose and disparity of the keyframes jointly, \cref{eq:dba}. Specifically, the objective is optimized over a local graph defined within a sliding window over the current frame.
\begin{equation}
    \mathop{\arg \min}_{\omega, d} \sum_{(i,j)\in E} \left\|\tilde{p}_{ij} - K\omega_j^{-1}(\omega_i(1/d_i)K^{-1}[p_i, 1]^T) \right\|^2_{\Sigma_{ij}}\enspace,
    \label{eq:dba}
\end{equation}
with $\tilde{p}_{ij} \in \mathbb{R}^{(W \times H \times 2) \times 1}$ being the flattened predicted pixel coordinates when the pixels $p_i \in \mathbb{R}^{(W \times H \times 2) \times 1}$ from keyframe $i$ are projected into keyframe $j$ using optical flow. Further, $K$ is the camera intrinsics, $\omega_j$ and $\omega_i$ the camera-to-world extrinsics for keyframes $j$ and $i$, $d_i$ the disparity of pixel $p_i$ and $\|\cdot\|_{\Sigma_{ij}}$ is the Mahalanobis distance with diagonal weighting matrix $\Sigma_{ij}$. Each weight denotes the confidence of the optical flow prediction for each pixel in $\tilde{p}_{ij}$. 
For clarity of the presentation, we omit homogeneous coordinates.
 
The second objective introduces monocular depth $D^{\text{mono}}$ as two additional data terms. The monocular depth $D^{\text{mono}}$ is predicted at runtime by a pretrained relative depth DPT model~\cite{eftekhar2021omnidata}. 
\begin{align}
    \label{eq:dba_wq}
    \mathop{\arg \min}_{d^{h}, \theta, \gamma}& \sum_{(i,j)\in E}\left\|\tilde{p}_{ij} - K\omega_j^{-1}(\omega_i(1/d_i^h)K^{-1}[p_i, 1]^T) \right\|^2_{\Sigma_{ij}} \\ 
    +& \alpha_1 \sum_{i\in V}\left\| d_i^h -  \left(\theta_i (1/D^{\text{mono}}_i) + \gamma_i\right)  \right\|^2  
    + \alpha_2 \sum_{i\in V} \left\| d_i^l -  \left(\theta_i (1/D^{\text{mono}}_i) + \gamma_i\right)  \right\|^2
    \enspace. \nonumber
\end{align}
Here, the optimizable parameters are the scales $\theta\in\mathbb{R}$, shifts $\gamma\in\mathbb{R}$ and a subset of the disparities $d^h$ classified as being high error (explained later). This is done since the monocular depth is only deemed useful where the multi-view disparity $d_i$ optimization is inaccurate. Furthermore, $\alpha_1 \!<\! \alpha_2$, which is done to ensure that the scales $\theta$ and shifts $\gamma$ are optimized with the preserved low error disparities $d^l$. The scale $\theta_i$ and shift $\gamma_i$ are initialized using least squares fitting 
\begin{equation}
    \{\theta_i,\,\gamma_i\} = \mathop{\arg \min}_{\theta,\gamma}\sum_{(u,v)}\Big( \big(\theta (1/D^{\text{mono}}_i)+\gamma\big) - d_i^l \Big)^2 \enspace.
    \label{eq:wq_least_square}
\end{equation}
\Cref{eq:dba} and \cref{eq:dba_wq} are optimized alternatingly to avoid the scale ambiguity encountered if $d$, $\theta$, $\gamma$ and $\omega$ are optimized jointly.  

Next, we describe how high and low error disparities are classified. For a given disparity map $d_i$ (separated into low and high error parts $\{d_i^l, d_i^h\}$) for frame $i$, we denote the corresponding depth $\tilde{D}_i = 1/d_i$. Pixel correspondences $(u,v)$ and $(\hat{u},\hat{v})$ between keyframes $i$ and $j$ respectively are established by warping $(u,v)$ into frame $j$ with depth $\tilde{D}_i$ as
\begin{equation}
        p_{i} = \omega_i\tilde{D}_i(u, v)K^{-1}[u, v, 1]^T  ,\qquad  [\hat{u},\hat{v}, 1]^T \propto K\omega_j^{-1}[p_{i},1]^T \enspace.
\end{equation}
The corresponding 3D point to $(\hat{u},\hat{v})$ is computed from the depth at $(\hat{u},\hat{v})$ as
\begin{equation}
        p_{j} = \omega_j\tilde{D}_j(\hat{u}, \hat{v})K^{-1}[\hat{u}, \hat{v}, 1]^T \enspace.
        \label{eq:back-project}
\end{equation}
If the L2 distance between $p_{i}$ and $p_{j}$ is smaller than a threshold, the depth $\tilde{D}_i(u, v)$ is consistent between $i$ and $j$. By looping over all keyframes except $i$, the global two-view consistency $n_{i}$ can be computed for frame $i$ as
\begin{equation}
    n_{i}(u,v) = \sum_{\substack{k\in \text{KFs},\\ k\neq i}}
    \mathds{1}\Big(
    \left\| p_{i} - p_{k} \right\|_2
    < \eta\cdot\text{average}(\tilde{D}_i)\Big) \enspace.
    \label{eq:two_view_consist}
\end{equation}
Here, $\mathds{1}(\cdot)$ is the indicator function and $\eta \in\mathbb{R}_{\geq 0}$ is a hyperparameter and $n_{i}$ is the total two-view consistency for pixel $(u,v)$ in keyframe $i$. $\tilde{D}_i(u, v)$ is valid if $n_{i}$ is larger than a threshold.

\boldparagraph{Loop Closure.} To mitigate scale and pose drift, we incorporate loop closure along with online global bundle adjustment (BA) in addition to local window frame tracking. Loop detection is achieved by calculating the mean optical flow magnitude between the current active keyframes (within the local window) and all previous keyframes. Two criteria are evaluated for each keyframe pair:
First, the optical flow must be below a specified threshold $\tau_{\text{loop}}$, ensuring sufficient co-visibility between the views.
Second, the time interval between the frames must exceed a predefined threshold $\tau_{t}$ to prevent the introduction of redundant edges into the graph. When both criteria are met, a unidirectional edge is added to the graph. During the loop closure optimization process, only the active keyframes and their connected loop nodes are optimized to keep the computational load manageable.



\boldparagraph{Global BA.} For the online global BA, a separate graph that includes all keyframes up to the present is constructed. Edges are introduced based on the temporal and spatial relationships between the keyframes, as outlined in \cite{zhang2023go}. Following the approach detailed in \cite{zhang2024glorie}, we execute online global BA every 20 keyframes. To maintain numerical stability, the scales of the disparities and poses are normalized prior to each global BA optimization. This normalization involves calculating the average disparity $\bar{d}$ across all keyframes and then adjusting the disparity to $d_{norm} = d/\bar{d}$ and the pose translation to $t_{norm} = \bar{d}t$.

\subsection{Deformable 3D Gaussian Scene Representation} \label{sec:3dgs_rep}
We adopt a 3D Gaussian Splatting representation~\cite{kerbl20233d} which deforms under DSPO or loop closure optimizations to achieve global consistency.
Thus, the scene is represented by a set $\mathcal{G} = \{g_i\}_{i=1}^N$ of 3D Gaussians. Each Gaussian primitive $g_i$, is parameterized by a covariance matrix \(\Sigma_i \in \mathbb{R}^{3 \times 3}\), a mean \(\bm{\mu}_i \in \mathbb{R}^3\), opacity \(o_i \in [0,1]\), and color $\mathbf{c}_i \in \mathbb{R}^3$. All attributes of each Gaussian are optimized through back-propagation. The density function of a single Gaussian is described as
\begin{equation}
    g_i(\mathbf{x}) = \exp\Big(-\frac{1}{2} (\mathbf{x}-\bm{\mu}_i)^\top \Sigma_i^{-1} (\mathbf{x}-\bm{\mu}_i) \Big) \enspace .
\end{equation}
Here, the spatial covariance $\Sigma_i$ defines an ellipsoid and is decomposed as $\Sigma_i = R_iS_iS^{T}_iR^{T}_i$, where $S_i = \mathrm{diag}(s_i) \in \mathbb{R}^{3\times3}$ is the spatial scale and $R_i \in \mathbb{R}^{3 \times 3}$ represents the rotation.

\boldparagraph{Rendering.} Rendering color and depth from $\mathcal{G}$, given a camera pose, involves first projecting (known as ``splatting'') 3D Gaussians onto the 2D image plane. This is done by projecting the covariance matrix $\Sigma$ and mean $\bm{\mu}$ as $\Sigma' = JR \Sigma R^T J^T$ and $\bm{\mu}' = K\omega^{-1}\bm{\mu}$, where $R$ is the rotation component of world-to-camera extrinsics $\omega^{-1}$ and $J$ is the Jacobian of the affine approximation of the projective transformation~\cite{zwicker2001surface}. The final pixel color $C$ and depth $D^r$ at pixel $\mathbf{x}'$ is computed by blending 3D Gaussian splats that overlap at a given pixel, sorted by their depth as
\begin{equation}
    C = \sum_{i \in \mathcal{N}} \mathbf{c}_i \alpha_i \prod_{j=1}^{i-1} (1 - \alpha_j) \qquad D^r = \sum_{i \in \mathcal{N}} \hat{d}_i \alpha_i \prod_{j=1}^{i-1} (1 - \alpha_j) \enspace ,
\end{equation}
where $\hat{d}_i$ is the z-axis depth of the center of the $i$-th 3D Gaussian and the final opacity $\alpha_i$ is the product of the opacity $o_i$ and the 2D Gaussian density as
\begin{equation}
    \alpha_i = o_i \exp\Big(-\frac{1}{2} (\mathbf{x}' - \bm{\mu}'_i)^\top \Sigma'^{-1}_i (\mathbf{x}' - \bm{\mu}'_i)\Big) \enspace.
\end{equation}
\boldparagraph{Map Initialization.} For every new keyframe, we adopt the RGBD strategy of MonoGS~\cite{matsuki2023gaussian} for adding new Gaussians to the unexplored scene space. 
As we do not have access to a depth sensor, we construct a proxy depth map $D$ by combining the inlier multi-view depth $\tilde{D}$ and the monocular depth $D^{\text{mono}}$ as
\begin{equation}
  D(u,v)= 
  \begin{cases}
    \tilde{D}(u,v) & \text{if $\tilde{D}(u,v)$ is valid}\\
    \theta D^{\text{mono}}(u,v)+\gamma & \text{otherwise}\enspace \\
  \end{cases}
\end{equation}
Here, $\theta$ and $\gamma$ are computed as in \cref{eq:wq_least_square} but using depth instead of disparity.

\boldparagraph{Keyframe Selection and Optimization.} 
Apart from the keyframe selection based on a mean optical flow threshold $\tau$, we additionally adopt the keyframe selection strategy from \cite{matsuki2023gaussian} to avoid mapping redundant frames. 

To optimize the 3D Gaussian parameters, we batch the parameter updates to a local window similar to \cite{matsuki2023gaussian} and apply a photometric and geometric loss to the proxy depth as well as a scale regularizer to avoid artifacts from elongated Gaussians. Inspired by \cite{matsuki2023gaussian}, we further use exposure compensation by optimizing an affine transformation for each keyframe. The final loss is
\begin{equation}
  \mathop{\min}_{\mathcal{G},\mathbf{a}, \mathbf{b}}\sum_{\substack{k\in \text{KFs}}} \frac{\lambda}{N_k} \lvert (a_kC_k+b_k) - C^{gt}_k \rvert_1 + \frac{1-\lambda}{N_k}\lvert D^r_k - D_k \rvert_1 + \frac{\lambda_{reg}}{|\mathcal{G}|}\sum_{i}^{|\mathcal{G}|}\lvert s_i - \tilde{s}_i \rvert_1 \enspace ,
  \label{eq:pix_warp_loss}
\end{equation}
where KFs contains the set of keyframes in the local window, $N_k$ is the number of pixels per keyframe, $\lambda$ and $\lambda_{reg}$ are hyperparameters, $\mathbf{a} = \{a_1,\dots,a_k, \dots\}$ and $\mathbf{b} = \{b_1,\dots, b_k, \dots\}$ are the parameters for the exposure compensation and $\tilde{s}$ is the mean scaling, repeated over the three dimensions.

\boldparagraph{Map Deformation.} 
Since our tracking framework is globally consistent, changes in the keyframe poses and proxy depth maps need to be accounted for in the 3D Gaussian map by a non-rigid deformation. Though the Gaussian means are directly optimized, one could in theory let the optimizer deform the map as refined poses and proxy depth maps are provided. We find, however, that in particular rendering is aided by actively deforming the 3D Gaussian map. We apply the deformation to all Gaussians which receive updated poses and depths before mapping. 

Each Gaussian $g_i$ is associated with a keyframe that anchored it to the map $\mathcal{G}$. Assume that a keyframe with camera-to-world pose $\omega$ and proxy depth $D$ is updated such that $\omega \rightarrow \omega'$ and $D \rightarrow D'$. We update the mean, scale and rotation of all Gaussians $g_i$ associated with the keyframe. Association is determined by what keyframe added the Gaussian to the scene. The mean $\bm{\mu}_i$ is projected into $\omega$ to find the pixel correspondence $(u,v)$. Since the Gaussians are not necessarily anchored on the surface, instead of re-anchoring the mean at $D'$, we opt to shift the mean by $D'(u,v) - D(u,v)$ along the optical axis. We update $R_i$ and $s_i$ accordingly as
\begin{equation}
    \bm{\mu}'_i = \!\bigg(\!1 + \frac{D'(u,v) - D(u,v)}{(\omega^{-1}\bm{\mu}_i)_z}\!\bigg)\omega'\omega^{-1}\bm{\mu}_i \enspace, 
    R_i' = R'R^{-1}R_i \enspace, 
    s'_i = \!\bigg(\!1 + \frac{D'(u,v) - D(u,v)}{(\omega^{-1}\bm{\mu}_i)_z}\!\bigg)s_i \enspace.
\end{equation}
Here, $(\cdot)_z$ denotes the z-axis depth. For Gaussians which project into pixels with missing depth or outside the viewing frustum, we \textit{only} rigidly deform them. After the final global BA optimization, we additionally deform the Gaussian map and perform a set of final refinements (see suppl. material).

\section{Experiments}
\label{sec:exp}

We first describe our experimental setup and then evaluate our method against state-of-the-art dense RGB and RGBD SLAM methods on Replica~\cite{straub2019replica} as well as the real world TUM-RGBD~\cite{Sturm2012ASystems} and the ScanNet~\cite{Dai2017ScanNet} datasets. 
For more experiments and details, we refer to the supplementary material.

\boldparagraph{Implementation Details.}
For the proxy depth, we use $\eta=0.01$ to filter points and use the condition $n_{c}\geq2$ to ensure multi-view consistency. For the mapping loss function, we use $\lambda=0.8$, $\lambda_{reg}=10.0$. We use 60 iterations during mapping. For tracking, we use $\alpha_1=0.01$ and $\alpha_2=0.1$ as weights for the DSPO layer. We use the flow threshold $\tau = 4.0$ on ScanNet, $\tau=3.0$ on TUM-RGBD and $\tau=2.25$ on Replica. The threshold for loop detection is $\tau_\text{loop} = 25.0$. The time interval threshold is $\tau_t = 20$. We conducted the experiments on a cluster with an NVIDIA A100 GPU.

\boldparagraph{Evaluation Metrics.}
For rendering we report PSNR, SSIM~\cite{wang2004image} and LPIPS~\cite{zhang2018unreasonable} on the rendered keyframe images against the sensor images. 
For reconstruction, we first extract the meshes with marching cubes~\cite{lorensen1987marching} as in~\cite{sandstrom2023point} and evaluate the meshes using accuracy $[cm]$, completion $[cm]$ and completion ratio $[\%]$ (threshold 5 cm) against the ground truth meshes. 
We also report the re-rendering depth L1 $[cm]$ metric to the ground truth sensor depth as in~\cite{Rosinol2022NeRF-SLAM:Fields}.
We use ATE RMSE $[cm]$~\cite{Sturm2012ASystems} to evaluate the estimated trajectory.

\boldparagraph{Datasets.} 
We use the RGBD trajectories from~\cite{Sucar2021IMAP:Real-Time} captured from the synthetic Replica dataset~\cite{straub2019replica}. We also test on real-world data using the
TUM-RGBD~\cite{Sturm2012ASystems} and the ScanNet~\cite{Dai2017ScanNet} datasets.

\boldparagraph{Baseline Methods.} We compare our method to numerous published and concurrent works on dense RGB and RGBD SLAM. Concurrent works are denoted with an asterix\textcolor{red}{$^*$}. The main baselines are GlORIE-SLAM~\cite{zhang2024glorie} and MonoGS~\cite{matsuki2023gaussian}.

\begin{table*}[tb]
    \centering    
    \scriptsize
    \setlength{\tabcolsep}{5.0pt}
    \resizebox{\columnwidth}{!}
    {
    \begin{tabular}{lccccccccc}
    \toprule
    Metric  & \makecell[c]{GO-\\SLAM~\cite{zhang2023go}} & \makecell[c]{NICER-\\SLAM~\cite{zhu2023nicer}} & \makecell[c]{MoD-\\SLAM$\textcolor{red}{^*}$~\cite{li2023dense}} &  \makecell[c]{Photo-\\SLAM~\cite{huang2023photo}}&  \makecell[c]{Mono-\\GS~\cite{matsuki2023gaussian}} & \makecell[c]{GlORIE-\\SLAM$\textcolor{red}{^*}$~\cite{zhang2024glorie}} & \makecell[c]{Q-SLAM\\$\textcolor{red}{^*}$~\cite{peng2024q}} & \makebox[0.07\linewidth]{\textbf{Ours}}\\
    \midrule
    PSNR$\uparrow$    & 22.13 & 25.41 & 27.31 & \nd 33.30 & \rd 31.22 & 31.04 & 32.49 & \fst 36.45 \\
    SSIM $\uparrow$   &  0.73 & 0.83  & 0.85  & \nd 0.93  & \rd 0.91 & \rd 0.91 & 0.89 & \fst 0.95\\
    LPIPS$\downarrow$ &  -   & \rd0.19  & -     & -     & 0.21 & \nd 0.12 & 0.17 & \fst 0.06\\  \midrule
    \makecell[l]{ATE RMSE}$\downarrow$& \nd0.39 & 1.88 & \fst 0.35 & \rd 1.09 & 14.54 &\fst 0.35 & - & \fst 0.35 \\ 
    \bottomrule
    \end{tabular}
    }
    \caption{
    \textbf{Rendering and Tracking Results on Replica~\cite{straub2019replica} for RGB-Methods}. Our method outperforms all methods on rendering and performs on par for tracking accuracy. Results are from~\cite{tosi2024nerfs} except ours (average over 8 scenes). Best results are highlighted as \colorbox{colorFst}{\bf first}, \colorbox{colorSnd}{second}, \colorbox{colorTrd}{third}.
    }
    \label{tab:render_replica}
\end{table*}

\begin{table*}[htb]
\centering
\scriptsize
\setlength{\tabcolsep}{12.4pt}
\begin{tabularx}{\linewidth}{llccccccc}
\toprule
Method & Metric & \texttt{0000} & \texttt{0059} & \texttt{0106} & \texttt{0169} & \texttt{0181} & \texttt{0207} & Avg.\\
\midrule
\multicolumn{9}{l}{\cellcolor[HTML]{EEEEEE}{\textit{RGB-D Input}}} \\ 





\multirow{3}{*}{\makecell[l]{SplaTaM~\cite{keetha2023splatam}}}
& PSNR$\uparrow$  & 19.33 & 19.27 & 17.73 & 21.97 & 16.76 & 19.80 & 19.14\\
& SSIM $\uparrow$ & 0.66 & \rd 0.79 & 0.69 & 0.78 & 0.68 & 0.70 & 0.72\\
& LPIPS$\downarrow$ &0.44 & 0.29 & 0.38 & 0.28 & 0.42 & \rd 0.34 & 0.36\\
[0.8pt] \hdashline \noalign{\vskip 1pt}

\multirow{3}{*}{\makecell[l]{MonoGS~\cite{matsuki2023gaussian}}} 
& PSNR$\uparrow$ & 18.70 &\rd 20.91 & 19.84 & 22.16 & \rd 22.01 & 18.90 & 20.42 \\
& SSIM $\uparrow$ & 0.71 & \rd 0.79 & 0.81 &0.78  & 0.82 & 0.75 & \rd 0.78\\
& LPIPS$\downarrow$ &0.48 & 0.32 & 0.32 & 0.34 &\rd 0.42 & 0.41 & 0.38 \\
[0.8pt] \hdashline \noalign{\vskip 1pt}

\multirow{3}{*}{\makecell[l]{Gaussian-\\SLAM~\cite{yugay2023gaussianslam}}}
& PSNR$\uparrow$  &\nd 28.54 & \nd 26.21 & \nd 26.26 & \nd 28.60 & \nd 27.79 & \nd 28.63 & \nd 27.67\\
& SSIM $\uparrow$ &\fst 0.93 & \fst 0.93 & \fst 0.93 & \fst 0.92 & \fst 0.92 & \fst 0.91 & \fst 0.92\\
& LPIPS$\downarrow$ &\rd 0.27 & \nd 0.21 & \nd 0.22 & \rd 0.23 &\nd  0.28 & \nd 0.29 & \nd 0.25\\

\midrule
\multicolumn{9}{l}{\cellcolor[HTML]{EEEEEE}{\textit{RGB Input}}} \\ 

\multirow{3}{*}{\makecell[l]{GO-\\SLAM~\cite{zhang2023go}}} 
& PSNR$\uparrow$ &15.74 & 13.15 & 14.58 & 14.49 & 15.72 & 15.37 & 14.84\\
& SSIM $\uparrow$ &0.42 & 0.32 & 0.46 & 0.42 & 0.53 & 0.39 & 0.42\\
& LPIPS$\downarrow$ &0.61 & 0.60 & 0.59 & 0.57 & 0.62 & 0.60 & 0.60\\
\midrule
%

\multirow{3}{*}{\makecell[l]{MonoGS~\cite{matsuki2023gaussian}}} 
& PSNR$\uparrow$ &16.91 & 19.15 & 18.57 & 20.21 & 19.51 & 18.37 & 18.79 \\
& SSIM $\uparrow$ & 0.62 & 0.69 & 0.74 & 0.74 & 0.75 & 0.70 & 0.71\\
& LPIPS$\downarrow$ &0.70 & 0.51 & 0.55 & 0.54 & 0.63 & 0.58 & 0.59 \\
[0.8pt] \hdashline \noalign{\vskip 1pt}

\multirow{3}{*}{\makecell[l]{GlORIE-\\SLAM$\textcolor{red}{^*}$~\cite{zhang2024glorie}}} 
& PSNR$\uparrow$ &\rd23.42 & 20.66 &\rd 20.41 &\rd 25.23 & 21.28 &\rd 23.68 &\rd 22.45\\
& SSIM $\uparrow$ & \nd 0.87 & \nd 0.87 & \rd 0.83 & \rd 0.84 & \nd 0.91 &\rd 0.76 & \nd 0.85 \\
& LPIPS$\downarrow$ &\nd 0.26 &\rd 0.31 &\rd 0.31 &\nd 0.21& 0.44 &\nd  0.29 &\rd 0.30\\
[0.8pt] \hdashline \noalign{\vskip 1pt}

\multirow{3}{*}{\makecell[l]{\textbf{\ours}\\ \textbf{(Ours)}}} 
    &PSNR$\uparrow$&\fst 28.68& \fst 27.69& \fst 27.70& \fst 31.14& \fst 31.15& \fst 30.49& \fst 29.48\\
     &SSIM $\uparrow$&\rd 0.83&\nd 0.87&\nd 0.86&\nd 0.87&\rd 0.84&\nd 0.84&\nd 0.85\\
     &LPIPS $\downarrow$& \fst 0.19& \fst 0.15& \fst 0.18& \fst 0.15& \fst 0.23& \fst 0.19& \fst 0.18\\
\bottomrule
\end{tabularx}

\caption{\textbf{Rendering Performance on ScanNet~\cite{Dai2017ScanNet}.} Our method performs even better or on par with all RGB-D methods. We take the numbers for SplaTaM and Gaussian-SLAM from~\cite{yugay2023gaussianslam}.}
\label{tab:render_scannet}
\end{table*}
\begin{table*}[htb]
\centering
\scriptsize
\setlength{\tabcolsep}{11.16pt}
\begin{tabularx}{\linewidth}{llcccccc}
\toprule
Method & Method  & \texttt{f1/desk} & \texttt{f2/xyz} & \texttt{f3/off}  & \texttt{f1/desk2} & \texttt{f1/room}& \textbf{Avg.}\\
\midrule
\multicolumn{8}{l}{\cellcolor[HTML]{EEEEEE}{\textit{RGB-D Input}}} \\ 
\multirow{3}{*}{\makecell[l]{SplaTaM~\cite{keetha2023splatam}}} 
    &PSNR$\uparrow$&\rd 22.00& 24.50& \rd 21.90& -& -& -\\
     &SSIM $\uparrow$&\nd 0.86&\fst 0.95&\nd 0.88& -& - & -\\
     &LPIPS $\downarrow$& \nd 0.23& \rd 0.10& \nd 0.20& -& -& -\\
[0.8pt] \hdashline \noalign{\vskip 1pt}
\multirow{3}{*}{\makecell[l]{Gaussian-\\SLAM~\cite{yugay2023gaussianslam}}} 
    &PSNR$\uparrow$&\nd  24.01& \rd 25.02& \fst 26.13& \nd 23.15 & \nd 22.98 & \nd 24.26\\
     &SSIM $\uparrow$&\fst  0.92&\nd 0.92&\fst 0.94& \nd 0.91 & \fst 0.89 & \fst 0.92 \\
     &LPIPS $\downarrow$& \fst 0.18&  0.19& \fst 0.14& \fst 0.20 & \fst 0.24 & \fst 0.19\\
\midrule
\multicolumn{8}{l}{\cellcolor[HTML]{EEEEEE}{\textit{RGB Input}}} \\ 
\multirow{3}{*}{\makecell[l]{Photo-SLAM~\cite{huang2023photo}}} 
    &PSNR$\uparrow$& 20.97& 21.07&  19.59& -& -& -\\
     &SSIM $\uparrow$&\nd 0.74& 0.73& 0.69& -& -& -\\
     &LPIPS $\downarrow$& \nd 0.23&  0.17& \rd 0.24& -& -& -\\
[0.8pt] \hdashline \noalign{\vskip 1pt}

\multirow{3}{*}{\makecell[l]{MonoGS~\cite{matsuki2023gaussian}}} 
    &PSNR$\uparrow$& 19.67 &  16.17 &  20.63 & \rd 19.16& 18.41 &  18.81\\
     &SSIM $\uparrow$& 0.73 & 0.72 & 0.77 &  0.66 &  0.64 & 0.70\\
     &LPIPS $\downarrow$& 0.33 & 0.31 & 0.34 & 0.48 & 0.51 &  \rd 0.39\\
[0.8pt] \hdashline \noalign{\vskip 1pt}

\multirow{3}{*}{\makecell[l]{GlORIE-\\SLAM$\textcolor{red}{^*}$~\cite{zhang2024glorie}}} 
    &PSNR$\uparrow$& 20.26&  \nd 25.62&  21.21 & 19.09& \rd 18.78&\rd  20.99\\
     &SSIM $\uparrow$& 0.79 &0.72 &0.72 &\fst 0.92 & \rd 0.73 & \rd 0.77\\
     &LPIPS $\downarrow$& \rd  0.31&  \nd 0.09& 0.32& \rd 0.38& \nd 0.38&\nd 0.30\\
[0.8pt] \hdashline \noalign{\vskip 1pt}

\multirow{3}{*}{\makecell[l]{\textbf{\ours}\\ \textbf{(Ours)}}}
    &PSNR$\uparrow$&\fst 25.61& \fst 29.53& \nd 26.05&  \fst 23.98& \fst 24.06 &\fst 25.85\\
     &SSIM $\uparrow$&\rd  0.84&\rd 0.90&\rd 0.84& \rd  0.81&\nd   0.80&\nd 0.84\\
     &LPIPS $\downarrow$& \fst 0.18&\fst 0.08& \nd 0.20& \nd 0.23& \fst 0.24&  \fst 0.19\\
\bottomrule
\end{tabularx}
\caption{\textbf{Rendering Performance on TUM-RGBD~\cite{Sturm2012ASystems}.} Our method performs competitively or better than RGB-D methods. For all RGB-D methods, we take the numbers from~\cite{yugay2023gaussianslam}.}
\label{tab:render_tum}
\end{table*}

\begin{figure}[ht]
\vspace{0em}
\centering
{
\setlength{\tabcolsep}{1pt}
\renewcommand{\arraystretch}{1}
\newcommand{\sz}{0.23}
\newcommand{\subsz}{0.2}
\begin{tabular}{ccccc}
\raisebox{0.6cm}{\rotatebox{90}{ \makecell{\texttt{Scene}\\\texttt{0000}}}}&
\includegraphics[width=\sz\linewidth]{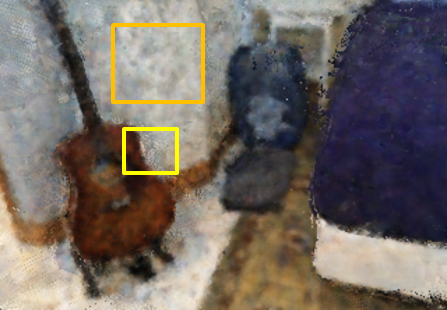} &
\includegraphics[width=\sz\linewidth]{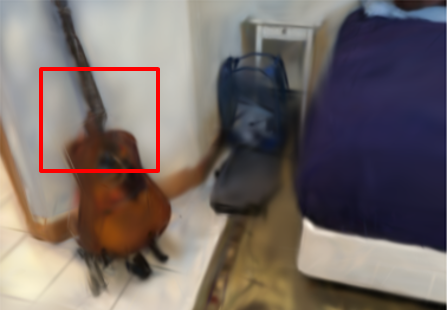} &
\includegraphics[width=\sz\linewidth]{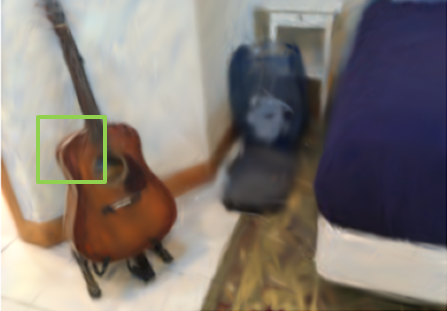} &
\includegraphics[width=\sz\linewidth]{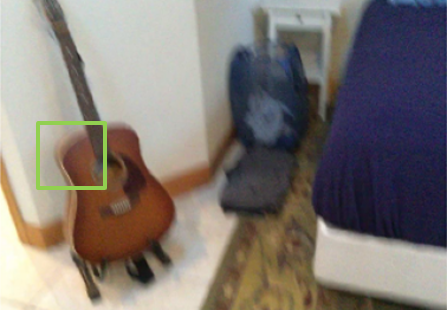} 
\\
\raisebox{0.75cm}{\rotatebox{90}{\textbf{\makecell{\texttt{Scene}\\\texttt{0054}}}}}&
\includegraphics[width=\sz\linewidth]{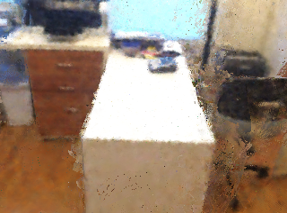} &
\includegraphics[width=\sz\linewidth]{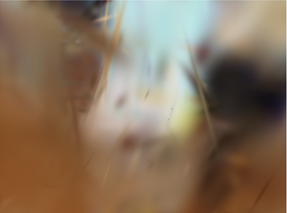} &
\includegraphics[width=\sz\linewidth]{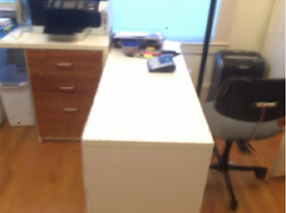} &
\includegraphics[width=\sz\linewidth]{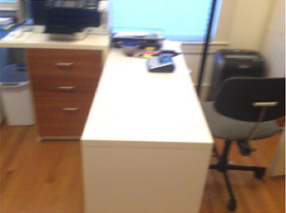} 
\\
\raisebox{0.6cm}{\rotatebox{90}{ \makecell{\texttt{fr3}\\\texttt{office}}}}&
\includegraphics[width=\sz\linewidth]{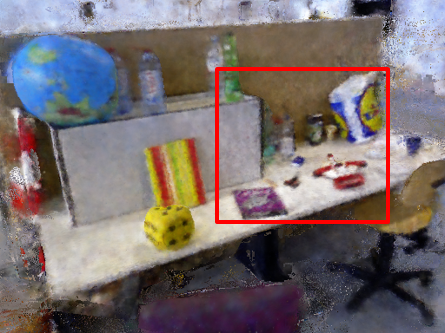} &
\includegraphics[width=\sz\linewidth]{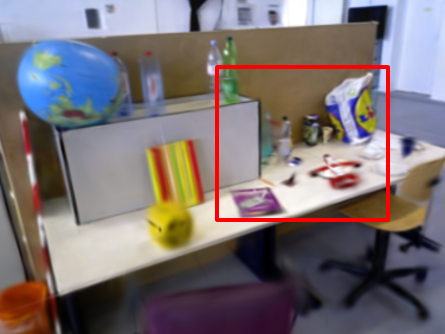} &
\includegraphics[width=\sz\linewidth]{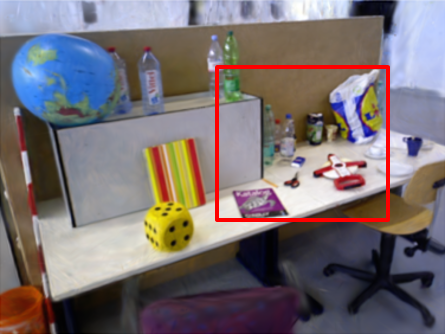} &
\includegraphics[width=\sz\linewidth]{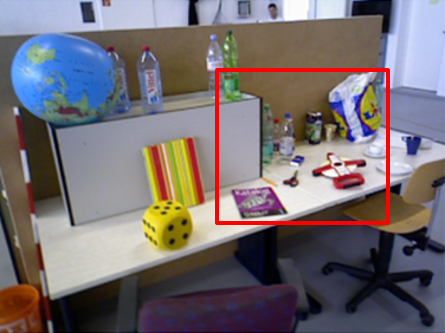} 
\\
&
\includegraphics[width=\sz\linewidth]{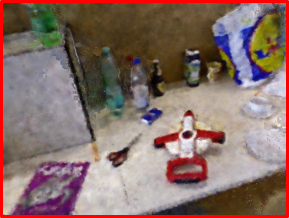} &
\includegraphics[width=\sz\linewidth]{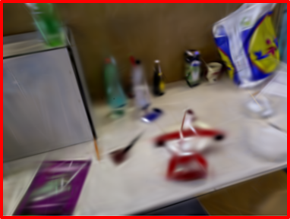} &
\includegraphics[width=\sz\linewidth]{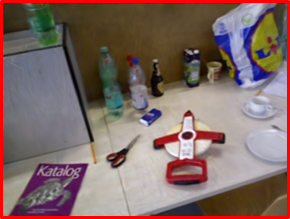} &
\includegraphics[width=\sz\linewidth]{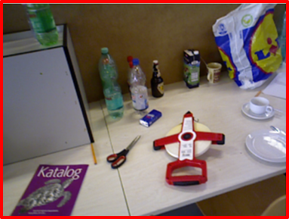} 
\\

\Large
& GlORIE-SLAM$\textcolor{red}{^*}$~\cite{zhang2024glorie}  & MonoGS~\cite{matsuki2023gaussian} & \ours (Ours)  & Ground Truth \\[-4pt]
\end{tabular}
}
\caption{\textbf{Rendering Results on ScanNet~\cite{Dai2017ScanNet} and TUM-RGBD~\cite{Sturm2012ASystems}.} Our method yields better rendering quality than GlORIE-SLAM and MonoGS. Top row: the orange box shows artifacts from GlORIE-SLAM, partly due to the depth guided volume rendering. The yellow box shows an area with redundant floating points. The red box shows a rendering distortion, likely from the large trajectory error. The green boxes show that our method fuses information from multiple views to avoid motion blur, present in the input. Fourth row: The rendering is from the pose of the red box in the third row.}
\label{fig:render_tum_scannet}
\vspace{0em}
\end{figure}

\boldparagraph{Rendering.}
In \cref{tab:render_replica}, we evaluate the rendering performance on Replica~\cite{straub2019replica} and find that our method performs superior among all baseline RGB-methods. \Cref{tab:render_scannet} and \cref{tab:render_tum} show the rendering accuracy on the ScanNet~\cite{Dai2017ScanNet} and TUM-RGBD~\cite{Sturm2012ASystems} datasets. In particular, we outperform existing RGB-only works with a clear margin, while even beating the currently best RGBD method, Gaussian-SLAM~\cite{yugay2023gaussianslam} on most metrics, despite the fact that we do not implement view-dependent rendering in the form of spherical harmonics. We attribute this to our deformable 3D Gaussian map, optimized with strong proxy depth along a globally consistent tracking backend. In \cref{fig:render_tum_scannet} and \cref{fig:teaser} we show renderings on the real-world ScanNet~\cite{Dai2017ScanNet} and TUM-RGBD~\cite{Sturm2012ASystems} datasets. Due to high tracking errors, MonoGS~\cite{matsuki2023gaussian} performs poorly on some scenes, yet fails to achieve the same fidelity as our method when the tracking error is low, as a result of the weak geometric constraints during optimization. Our method avoids the artifacts produced by GlORIE-SLAM~\cite{zhang2024glorie} and yields high quality renderings. 


\begin{table*}[t]
    \centering
    \scriptsize
    \setlength{\tabcolsep}{2.0pt}
    \resizebox{\columnwidth}{!}
    {
    \begin{tabular}{lcccccccccc}
    \toprule
    Metrics  & \makecell[l]{NeRF-\\SLAM~\cite{tosi2024nerfs}} & \makecell[l]{DIM-\\SLAM~\cite{li2023dense}} & \makecell[l]{GO-\\SLAM~\cite{zhang2023go}} &  \makecell[l]{NICER-\\SLAM~\cite{zhu2023nicer}}&  \makecell[l]{HI-\\SLAM~\cite{zhang2023hi}} & \makecell[l]{MoD-\\SLAM$\textcolor{red}{^*}$\!~\cite{zhou2024modslam}} & \makecell[l]{GlORIE-\\SLAM$\textcolor{red}{^*}$\!~\cite{zhang2024glorie}} & \makecell[l]{Mono-\\GS\!~\cite{matsuki2023gaussian}} & \makecell[l]{Q-SLAM\\$\textcolor{red}{^*}$\!~\cite{peng2024q}} & \makebox[0.1\linewidth]{\textbf{Ours}}\\
    \midrule
    \makecell[l]{Render Depth L1}$\downarrow$& \rd 4.49 & - & - & - & - & - & - & 27.24 & \nd 2.76 & \fst 2.41\\
    Accuracy $\downarrow$             & - & 4.03 & 3.81 & 3.65 & 3.62 & \nd 2.48 & \rd 2.96 & 30.61 & -& \fst 2.43 \\
    Completion $\downarrow$           & - & 4.20 & 4.79 & \rd 4.16 & 4.59 & - & \nd 3.95 & 12.19 & -& \fst 3.64 \\ 
    Comp. Rat. $\uparrow$             & - & 79.60 & 78.00 & 79.37& \rd 80.60 & - & \nd83.72  & 40.53 & -& \fst 84.69 \\
    \bottomrule
    \end{tabular}
    }
    \caption{
    \textbf{Reconstruction Results on Replica~\cite{straub2019replica} for RGB-Methods.} Our method outperforms existing works on all metrics. Results are averaged over 8 scenes.
    \label{tab:recon_replica}}
\end{table*}

\begin{figure}[ht]
\vspace{0em}
\centering
{
\setlength{\tabcolsep}{1pt}
\renewcommand{\arraystretch}{1}
\newcommand{\sz}{0.23}
\newcommand{\subsz}{0.2}
\begin{tabular}{ccccc}
\raisebox{0.4cm}{\rotatebox{90}{\makecell{\texttt{Office 0}}}}&
\includegraphics[width=\sz\linewidth]{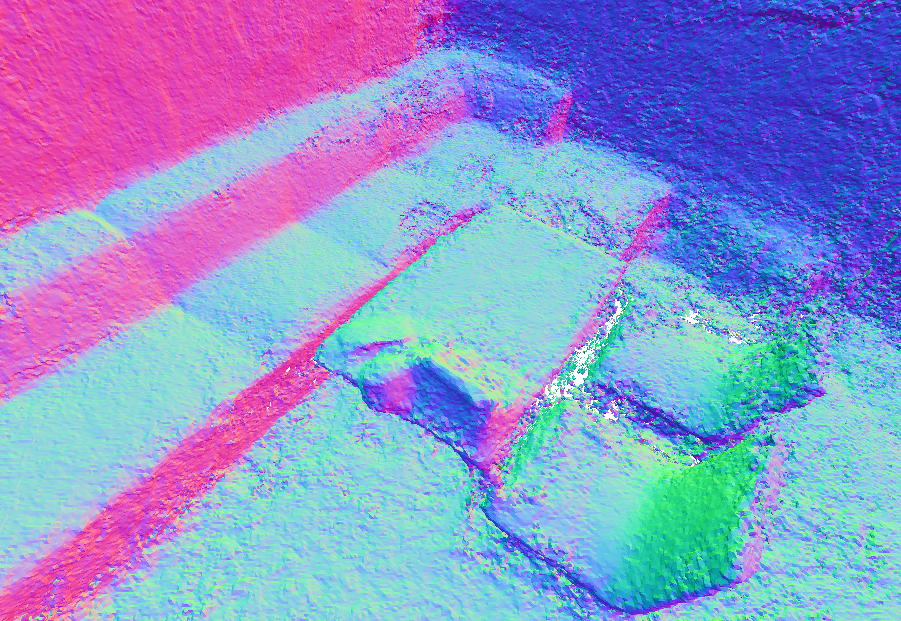} &
\includegraphics[width=\sz\linewidth]{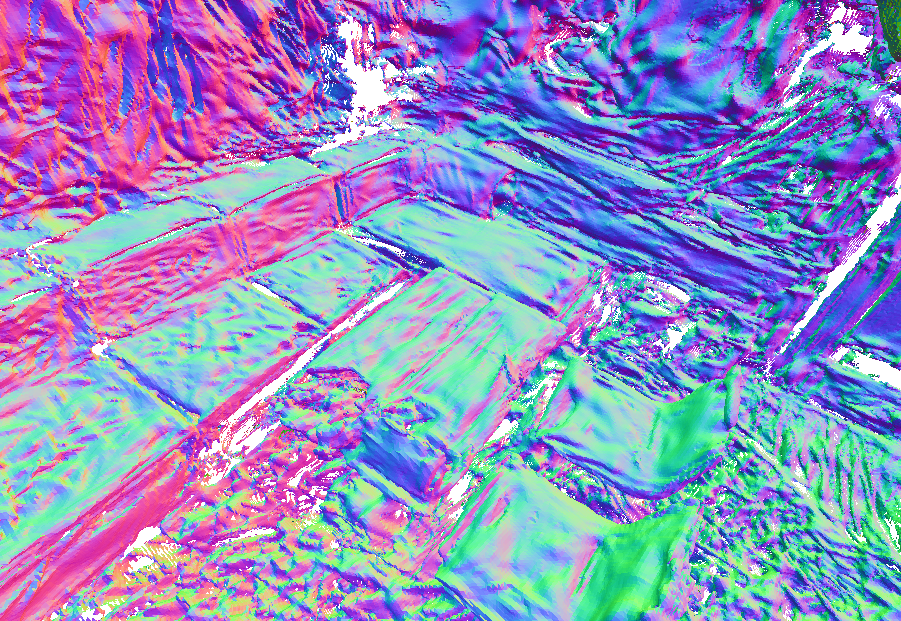} &
\includegraphics[width=\sz\linewidth]{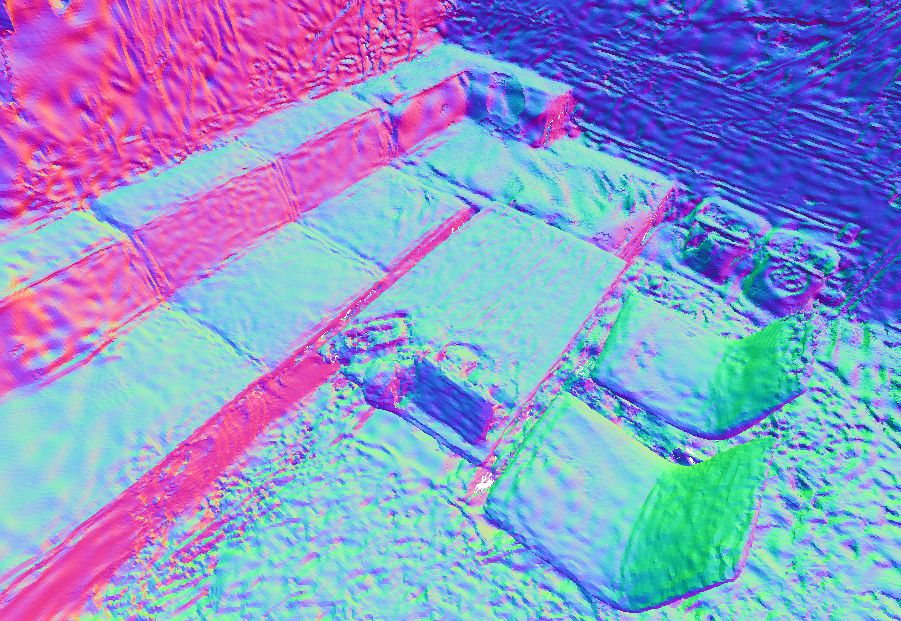} &
\includegraphics[width=\sz\linewidth]{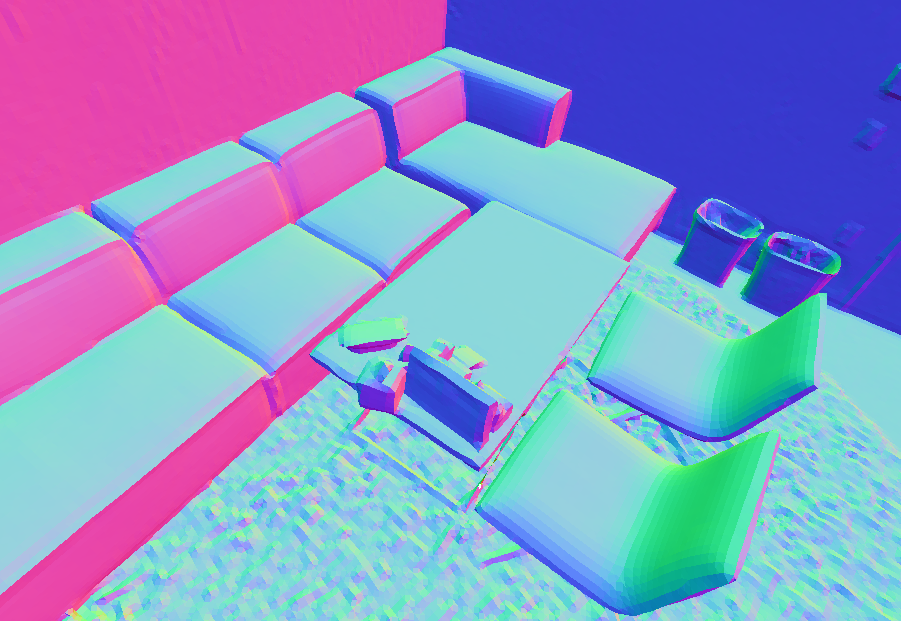} 
\\
\raisebox{0.25cm}{\rotatebox{90}{ \makecell{\texttt{Office 4}}}}&
\includegraphics[width=\sz\linewidth]{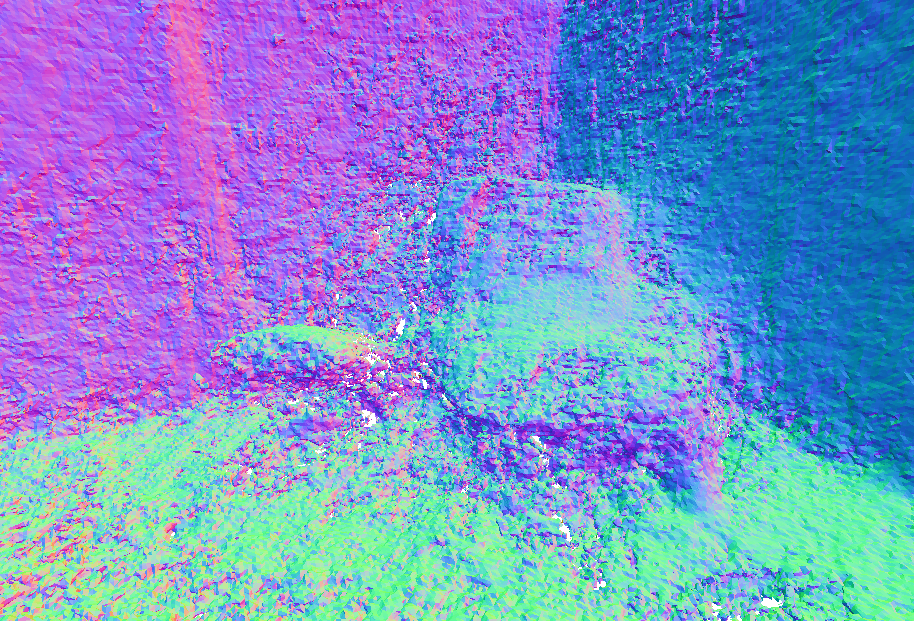} &
\includegraphics[width=\sz\linewidth]{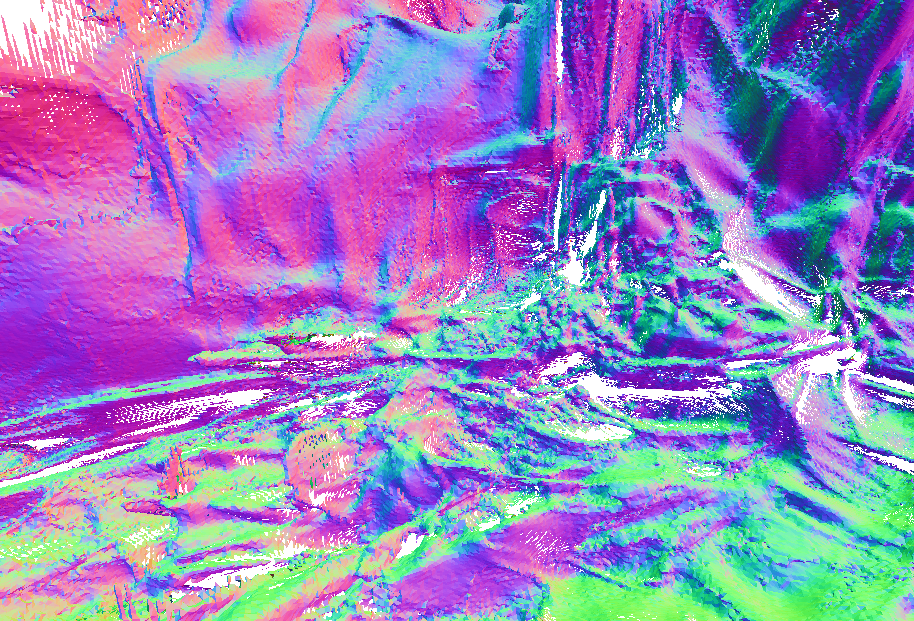} &
\includegraphics[width=\sz\linewidth]{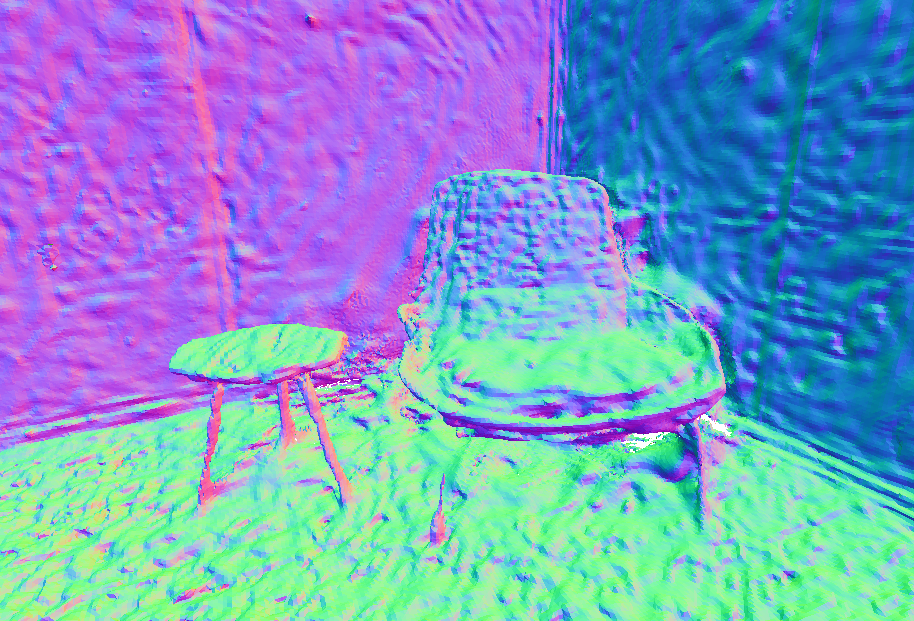} &
\includegraphics[width=\sz\linewidth]{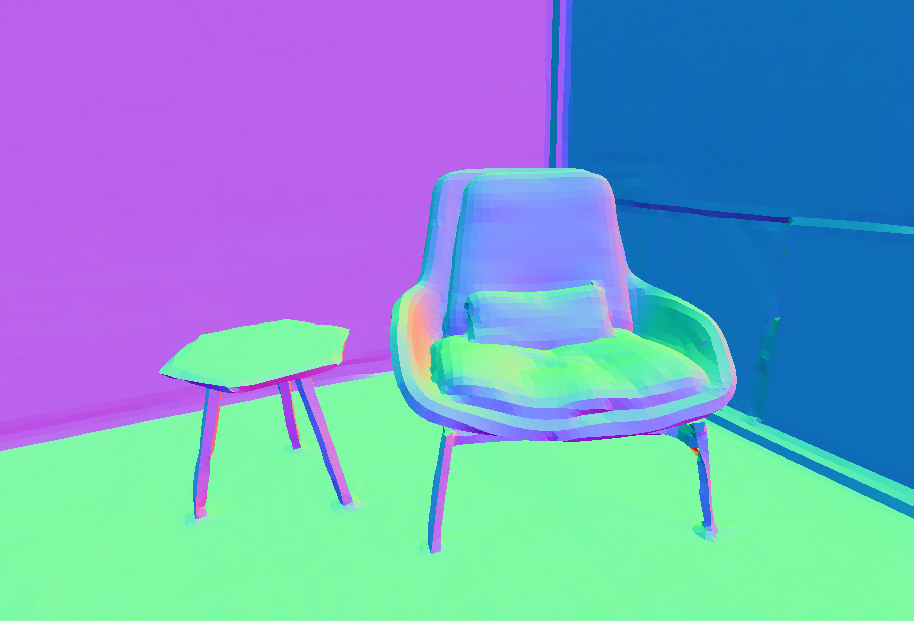} 
\\
\raisebox{0.5cm}{\rotatebox{90}{ \makecell{\texttt{Room 0}}}}&
\includegraphics[width=\sz\linewidth]{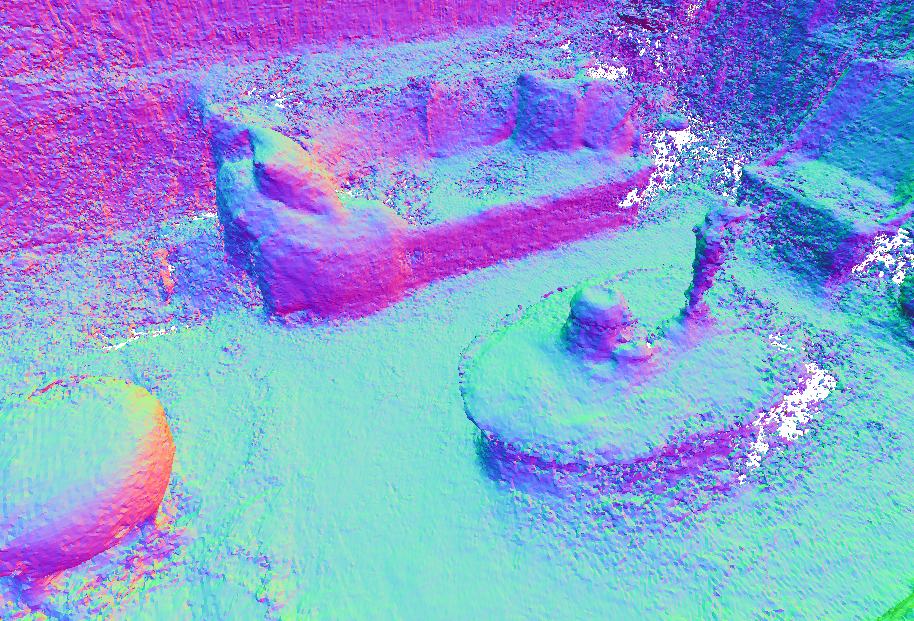} &
\includegraphics[width=\sz\linewidth]{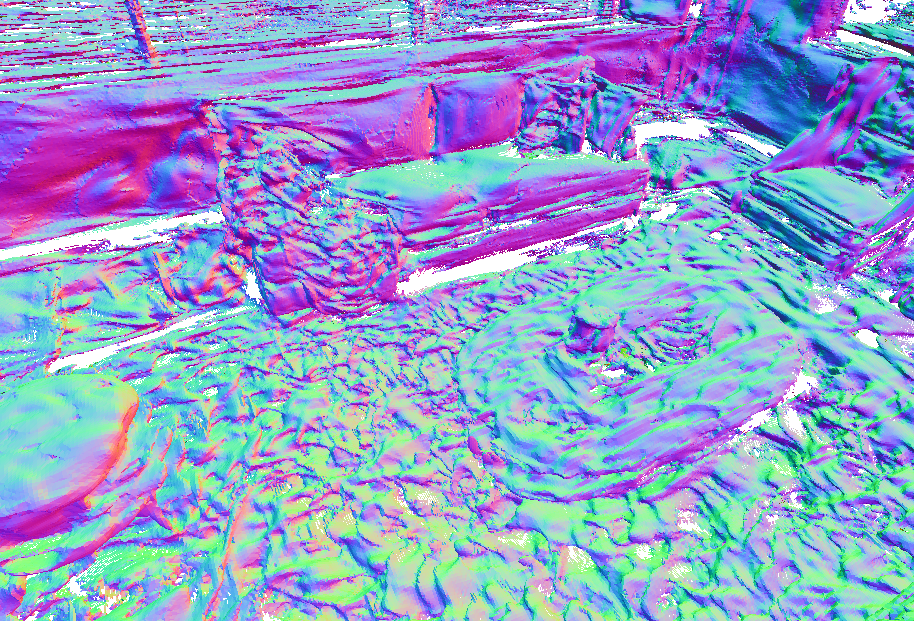} &
\includegraphics[width=\sz\linewidth]{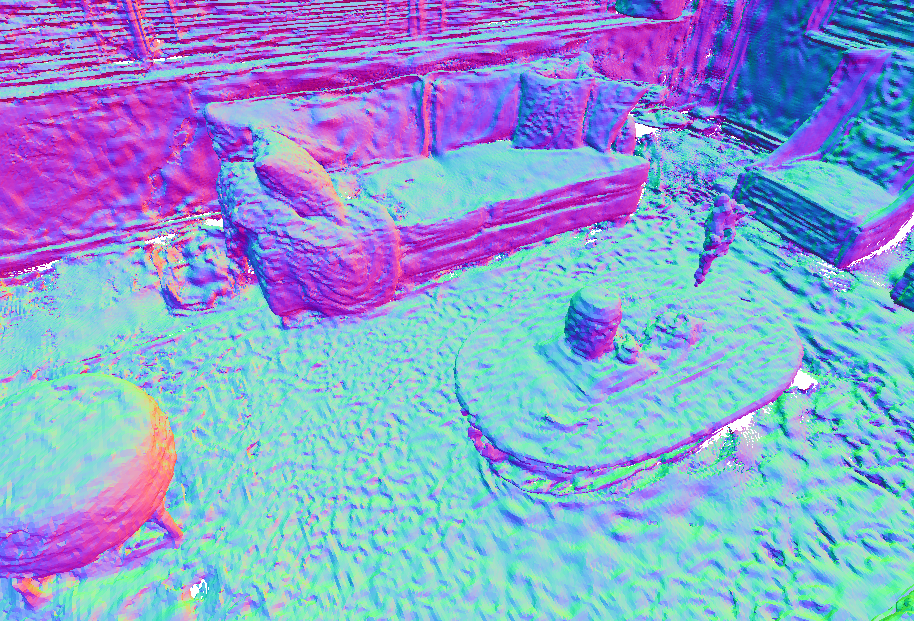} &
\includegraphics[width=\sz\linewidth]{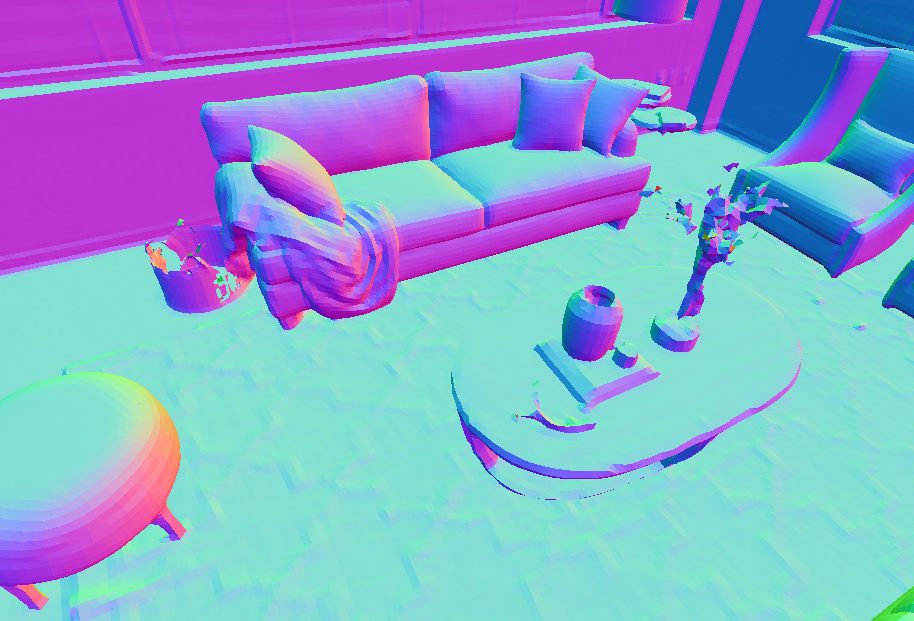} 
\\

\Large
& GlORIE-SLAM$\textcolor{red}{^*}$~\cite{zhang2024glorie}  & MonoGS~\cite{matsuki2023gaussian} & \ours (Ours)  & Ground Truth \\
\end{tabular}
}
\caption{\textbf{Reconstruction Results on Replica~\cite{straub2019replica} on Normal Shaded Meshes.} Our method achieves higher geometric accuracy compared to existing works. In particular, GlORIE-SLAM suffers from floating point artifacts (\eg second row) where our method reconstructs even the individual legs of the table. MonoGS suffers significantly from a lack of proxy depth, despite multiview optimization.
\label{fig:reconstruction_replica}}
\vspace{0em}
\end{figure}

\boldparagraph{Reconstruction.}
We show quantitative and qualitative results on the Replica~\cite{straub2019replica} dataset in \cref{tab:recon_replica} and \cref{fig:reconstruction_replica} respectively. 
Our method achieves the best performance on all metrics. Qualitatively, we show normal shaded meshes from different viewpoints. Our method can reconstruct finer details than existing works, especially around thin structures (\eg second row), where our strong proxy depth coupled with the 3D Gaussian map representation yields superior depth rendering, which directly influences the mesh quality. In contrast, \eg GlORIE-SLAM~\cite{zhang2024glorie} uses depth guided volume rendering, which is sensitive to input depth noise, resulting in inconistent depth rendering with floating artifacts. MonoGS~\cite{matsuki2023gaussian} suffers significantly from the lack of proxy depth, visible in all scenes. \Cref{fig:teaser} shows depth rendering on the real-world TUM-RGBD~\cite{Sturm2012ASystems} \texttt{room} scene. We compute the average depth L1 error over all keyframes, achieving 15.05 cm, beating existing works.

\begin{table}[htb]
    \centering
    \scriptsize
    \setlength{\tabcolsep}{14.8pt}
    \begin{tabularx}{\linewidth}{ccccccc}
    \toprule
    %
    \makecell[c]{Mono \\ Depth}& \makecell[c]{Multiview\\Depth} &  \makecell[c]{Multiview \\Filtering}& \makecell[c]{PSNR\\ \,[dB] $\uparrow$} &\makecell[c]{Acc.\\ \,[cm] $\downarrow$} & \makecell[c]{Comp.\\ \,[cm] $\downarrow$}& \makecell[c]{Comp. Ratio\\ \,[cm] $\uparrow$}\\
    \midrule
    \greencheck & \redx & \redx & 36.02 & \rd 3.62 & 4.08 \rd & \nd 81.16 \\
     \redx & \greencheck & \greencheck & \rd 36.17 & \nd 2.64 & 4.73 &  80.12\\
     \redx & \greencheck & \redx & \nd 36.21 & 18.71 & \nd 4.06 & \rd 80.29\\
     \greencheck & \greencheck & \greencheck & \fst 36.45 & \fst 2.43 & \fst 3.64 & \fst 84.69\\
    \bottomrule
    \end{tabularx}
    \vspace{1pt}
    \caption{
    \textbf{Ablation Study on Replica~\cite{straub2019replica}.} We show that the combination of filtered multiview depth completed with monocular depth yields the best performance on all metrics. Mono Depth refers to $D^{mono}$, Multiview Depth refers to $\tilde{D}$ and Multiview Filtering means enabling \cref{eq:two_view_consist}. All results are averaged over 8 scenes.
    }
    \label{tab:ablation}
\end{table}

\boldparagraph{Ablation Study.}
In \cref{tab:ablation}, we conduct a set of ablation studies related to our method, by enabling and disabling certain parts. We find that the combination of filtered multiview depth completed with monocular depth yields the best performance in terms of rendering and reconstruction metrics.

\begin{table}[!htb]
    \vspace{-10pt}  
    \centering
    \scriptsize
    \setlength{\tabcolsep}{11.5pt}
    \begin{tabular}{lccccc}
    \toprule
       & GO-SLAM~\cite{zhang2023go} & SplaTAM~\cite{keetha2023splatam} & GlORIE-SLAM$\textcolor{red}{^*}$~\cite{zhang2024glorie} & MonoGS~\cite{matsuki2023gaussian} &Ours\\
    \midrule
    GPU Usage [GiB] & 18.50 & 18.54 & \nd 15.22 & \fst 14.62 & \rd 17.57\\[0.8pt] \hdashline \noalign{\vskip 1pt}
    Map Size [MB] & - & - & \rd 114.0 & \nd 6.8 & \fst 6.5\\[0.8pt] \hdashline \noalign{\vskip 1pt}
    Avg. FPS & \fst 8.36 & 0.14 & 0.23 & \rd 0.32 & \nd 1.24\\
    \bottomrule
    \end{tabular}
    \vspace{2pt}
    \caption{
    \textbf{Memory and Running Time Evaluation on Replica~\cite{straub2019replica} \texttt{room0}.} Our peak memory usage and runtime are comparable to existing works. We take the numbers from~\cite{tosi2024nerfs} except for ours and MonoGS and we add the Map Size, which denotes the size of the final 3D representation. GPU Usage denotes the peak usage during runtime. All methods are evaluated on an NVIDIA RTX 3090 GPU using single threading for fairness. 
    }
    \label{tab:mem_and_time}
\end{table}

\boldparagraph{Memory and Runtime.}
In \cref{tab:mem_and_time}, we evaluate the peak GPU memory usage, map size and runtime of our method. We achieve a comparable GPU memory usage with GO-SLAM~\cite{zhang2023go} and SplaTaM~\cite{keetha2023splatam}. Our map size is similar to MonoGS~\cite{matsuki2023gaussian} and much smaller than GlORIE-SLAM, which does not prune redundant neural points. In \cref{fig:teaser} we also show similar map size to MonoGS on the real-world TUM-RGBD~\cite{Sturm2012ASystems} \texttt{room} scene.
Regarding runtime, we are faster than SplaTaM and GlORIE-SLAM and comparable to MonoGS. GO-SLAM has the fastest runtime, but as shown in \cref{tab:render_replica} and \cref{tab:recon_replica}, it sacrifices rendering and reconstruction quality for speed.  

\boldparagraph{Limitations.} 
We currently do not model the appearance with spherical harmonics, since it only yields a marginal gains in rendering accuracy, while requiring more memory. It is is straightforward to add.  
We only make use of globally optimized frame-to-frame tracking, which fails to leverage frame-to-model queues from the 3D Gaussian map. Another limitation is that our construction of the final proxy depth $D$ is quite simple and does not fuse the monocular and keyframe depths in an informed manner, \eg using normal consistency. Finally, as future work, it is interesting to study how surface regularization can be enforced via \eg quadric surface elements as in \cite{peng2024q}. 

\section{Conclusion}
\label{sec:conclusion}
We proposed \ours, a dense RGB-only SLAM system which uses a deformable 3D Gaussian map for mapping and globally optimized frame-to-frame tracking via optical flow.
Importantly, the inclusion of monocular depth into the tracking loop, to refine the scale and to correct the erroneous keyframe depth predictions, leads to better rendering and mapping. 
By using the monocular depth for completion, mapping is further improved. 
Our experiments demonstrate that \ours outperforms existing solutions regarding reconstruction and rendering accuracy while being on par or better with respect to tracking as well as runtime and memory usage.


\clearpage

\appendix

\newcommand{\supplementarytitle}{
  \hrule height 4pt
  \vskip 0.25in
  \vskip -\parskip%
  \begin{centering}
  {\LARGE\bf Supplementary Material \\ \ours: Globally Optimized RGB-only SLAM with 3D Gaussians \par}
  \end{centering}
  \vskip 0.29in
  \vskip -\parskip
  \hrule height 1pt
  \vskip 0.09in%
}

\newcommand{\supplementaryauthor}{
  \begin{center}
    \renewcommand{\thefootnote}{\fnsymbol{footnote}}
    \begin{tabular}{cccc}
      \textbf{Erik Sandström\footnotemark[1]} & \textbf{Keisuke Tateno} & \textbf{Michael Oechsle} & \textbf{Michael Niemeyer} \\
      Google & Google & Google & Google \\
      ETH Zürich & & & \\
      \\
      \multicolumn{4}{c}{
        \begin{tabular}{ccc}
          \textbf{Luc Van Gool} & \textbf{Martin R. Oswald} & \textbf{Federico Tombari} \\
          ETH Zürich & ETH Zürich & Google \\
          INSAIT & University of Amsterdam & TU München \\
        \end{tabular}
      }
    \end{tabular}
    \footnotetext[1]{This work was conducted during an internship at Google.}
    \renewcommand{\thefootnote}{\arabic{footnote}}
  \end{center}
  \vskip 0.3in minus 0.1in
}

\supplementarytitle
\supplementaryauthor

\renewcommand{\thefigure}{S\arabic{figure}}
\renewcommand{\thetable}{S\arabic{table}}
\setcounter{figure}{0}
\setcounter{table}{0}
\setcounter{section}{0}

%
\setcounter{table}{7}
\setcounter{equation}{15}
\setcounter{figure}{5}

\label{sec:supp}


This supplementary material accompanies the main paper and provides more details on the methodology and additional experimental results.


\section{Method}
We describe further details about our method that were left out from the main paper.

\boldparagraph{Comparison to Existing Works.} To further clarify the differences between our method and existing 3DGS SLAM works, we classify each method in \cref{tab:methods} based on important characteristics. 
It shows that our work is the first to include loop closure, proxy depth, RGB-only and online 3D Gaussian deformations.

\begin{table}[htb]
    \centering
    \scriptsize
    \setlength{\tabcolsep}{20.5pt}
    \begin{tabularx}{\linewidth}{lcccc}
    \toprule
    & \makecell[c]{RGB-only}& \makecell[c]{Loop\\Closure} &  \makecell[c]{Proxy\\Depth}& \makecell[c]{Online 3DGS\\Deformations} \\
    \midrule
    GS-SLAM~\cite{yan2023gs} & \redx & \redx & \greencheck & \redx  \\
    Gaussian-SLAM~\cite{yugay2023gaussianslam} & \redx & \redx & \greencheck & \redx  \\
    SplaTaM~\cite{keetha2023splatam} & \redx & \redx & \greencheck & \redx \\
    MonoGS~\cite{matsuki2023gaussian} & \greencheck & \redx & \redx & \redx \\
    Photo-SLAM~\cite{huang2023photo} & \greencheck & \greencheck & \redx & \redx  \\
    \ours (ours) & \greencheck & \greencheck & \greencheck& \greencheck  \\     
    \bottomrule
    \end{tabularx}
    \caption{
    \textbf{Method Classification.} We show that our method is the first to combine 3D Gaussian SLAM with loop closure, proxy depth and online 3D Gaussian map deformations in an RGB-only SLAM system.
    }
    \label{tab:methods}
\end{table}





\boldparagraph{Map Initialization.}
With map initialization, we refer to the process of anchoring new Gaussians during scene exploration. For every new keyframe to be mapped, we adopt the strategy that MonoGS~\cite{matsuki2023gaussian} uses in pure RGBD mode. It works by unprojecting the depth reading per pixel to 3D and then downsampling this point cloud by a factor $\theta$. New Gaussians are then assigned their means as the point cloud. The rotations are initialized to identity, the opacity to 0.5 and the scales are initialized related to their distance to the nearest neighbor point in the point cloud.

\boldparagraph{Keyframe Selection and Local Windowing.}
As mentioned in the main paper, we adopt the keyframe selection strategy from MonoGS~\cite{matsuki2023gaussian}. We describe this strategy in the following. 

Keyframes are selected based on the covisibility of the Gaussians. Between two keyframes \(i\) and \(j\), the covisibility is defined using the Intersection over Union (IOU) and Overlap Coefficient (OC):
\begin{equation}
    \text{IOU}_{\text{cov}}(i, j) = \frac{|\mathcal{G}_v^i \cap \mathcal{G}_v^j|}{|\mathcal{G}_v^i \cup \mathcal{G}_v^j|}, \label{eq:iou_cov}
\end{equation}
\begin{equation}
    \text{OC}_{\text{cov}}(i, j) = \frac{|\mathcal{G}_v^i \cap \mathcal{G}_v^j|}{\min(|\mathcal{G}_v^i|, |\mathcal{G}_v^j|)}, \label{eq:oc_cov}
\end{equation}
where \(\mathcal{G}_v^i\) are the Gaussians visible in keyframe \(i\), based on the following definition of visibility. A Gaussian is seen as visible from a camera pose if it is used in the rasterization pipeline when rendering and if the accumulated transmittance $\prod_{j=1}^{i-1} (1 - \alpha_j)$ has not yet reached 0.5.

A keyframe \(i\) is added to the keyframe window \(\text{KFs}\) if, given the last keyframe \(j\), \(\text{IOU}_{\text{cov}}(i, j) < k_{\text{fcov}}\) or if the relative translation \(t_{ij} > k_{\text{fm}} \hat{D}_i\), where \(\hat{D}_i\) is the median depth of frame \(i\). For Replica, \(k_{\text{fcov}} = 0.95\), \(k_{\text{fm}} = 0.04\) and for TUM and ScanNet, \(k_{\text{fcov}} = 0.90\), \(k_{\text{fm}} = 0.08\). The registered keyframe \(j\) in \(\text{KFs}\) is removed if \(\text{OC}_{\text{cov}}(i, j) < k_{\text{fc}}\), where keyframe \(i\) is the latest added keyframe. For all datasets, the cutoff is set to \(k_{\text{fc}} = 0.3\). The size of the keyframe window is set to \(|\text{KFs}| = 10\) for Replica and \(|\text{KFs}| = 8\) for TUM and ScanNet.

\boldparagraph{Pruning and Densification}
We also follow \cite{matsuki2023gaussian} when it comes to Gaussian pruning and densification. Pruning is done based on the visibility: if new Gaussians inserted
within the last 3 keyframes are not visible by at least 3
other frames in the keyframe window KFs, they are removed. Visibility-based pruning is only done when the keyframe window KFs is full. Additionally, every 150 mapping iterations, Gaussians with opacity lower than 0.7 are removed globally. Also Gaussians which project in 2D with a too large scale are removed. Densification is done as in \cite{kerbl20233d}, also at an interval of every 150 mapping iterations.

\boldparagraph{Final Refinement.}
Similar to GlORIE-SLAM~\cite{zhang2024glorie}, which performs a final refinement after the last final global BA at the end of the trajectory, we also perform a few refinement iterations after the last final global BA. Also MonoGS~\cite{matsuki2023gaussian} performs a set of final iterations at the end of the SLAM trajectory to refine the colors.

Our refinement strategy is straight forward. We disable pruning and densification of the Gaussians and perform a set of optimization iterations $\beta$ using the same loss function as in the main paper, but only sampling random single frames per iteration.

\boldparagraph{Differences to GlORIE-SLAM.} 
We briefly discuss some differences to GlORIE-SLAM \cite{zhang2024glorie} not covered in the main paper. GlORIE-SLAM uses an additional point cloud called $P_d$ consisting of all inlier mullti-view depth maps unprojected into a point cloud. We found that this is not needed and it saves memory and compute to not use it. GlORIE-SLAM also re-anchors the neural points at the depth reading. We do not do this as the Gaussians do not necessarily lie on the surface exactly. Finally, GlORIE-SLAM requires input depth to guide the sampling of points to render color and depth. If the depth is noisy or if the map is used for tracking (\ie frame-to-model tracking), the depth guiding strategy is not favorable as it leads to artifacts when sampling the wrong points (when noisy depth is encountered) and to a much smaller basin of convergence when tracking (because the rendering is conditioned on the current view point). With 3D Gaussians, we can avoid depth guidance during rendering.

\section{More Experiments}
To accompany the evaluations provided in the main paper, we provide further experiments in this section. 

\boldparagraph{Implementation Details.}
As the point cloud downsampling factor, we use $\theta = 32$ for all frames but the first frame where $\theta = 16$ is used. We use $\beta = 2000$, the number of iterations for the final refinement optimization, on the Replica dataset and $\beta = 26000$ on the TUM-RGBD~\cite{Sturm2012ASystems} and ScanNet~\cite{Dai2017ScanNet} datasets (same as MonoGS~\cite{matsuki2023gaussian}). We benchmark the runtime on an AMD Ryzen Threadripper Pro 3945WX 12-Cores with an NVIDIA GeForce RTX 3090 Ti with 24 GB of memory. For the remaining hyperparameters, we refer to MonoGS \cite{matsuki2023gaussian} for the Gaussian mapping and GlORIE-SLAM for tracking \cite{zhang2024glorie}. 

\boldparagraph{A Note on Rendering and Runtime with MonoGS.} By default, MonoGS~\cite{matsuki2023gaussian} does not evaluate the rendering error on the mapped keyframes nor implement the exposure compensation during rendering evaluation. To compare our results fairly to MonoGS, we implement these details and run the experiments with these settings enabled. Further, we report the runtime for MonoGS using a single process (same as us) compared to the reported number in the paper, which was using multiple processes at once.

\boldparagraph{A Note on Gaussian Deformation with Photo-SLAM.} Though not fully clear from reading the paper, after discussing with the authors of Photo-SLAM~\cite{huang2023photo}, we find that they do, in fact, not deform the Gaussians as a result of global BA or loop closure. They found this to be unstable in their experiments. This suggests that our deformation strategy is non-trivial.

\boldparagraph{Justification of Monocular Depth Estimator.} 
There are already numerous monocular depth estimators, but most of them are limited by speed, memory or quality. We use Omnidata \cite{eftekhar2021omnidata} since empirically we found it still provides the best trade-off between output performance and runtime. We also tested our system with Depth Anything \cite{yang2024depth}, but found that it was marginally worse in terms of the final reconstructed mesh accuracy.

\subsection{Tracking on ScanNet and TUM-RGBD}
We do not put the results on tracking for ScanNet and TUM-RGBD since we use the tracking framework from GlORIE-SLAM~\cite{zhang2024glorie}, but we provide the numbers here. \Cref{tab:track_scannet} and \cref{tab:track_tum_rgbd} show the tracking accuracy of the estimated trajectory on  ScanNet~\cite{Dai2017ScanNet} and TUM-RGBD~\cite{Sturm2012ASystems} respectively. Our method shows competitive results in every single scene and gives the best average value among the RGB and RGB-D methods.

\begin{table}[htb]
    \centering
    \scriptsize
    \setlength{\tabcolsep}{5.4pt}
    \begin{tabular}{lcccccccccc}
    \toprule
    Method  & \texttt{0000} & \texttt{0059} & \texttt{0106} & \texttt{0169}& \texttt{0181} & \texttt{0207} & \textbf{Avg.-6} & \texttt{0054} & \texttt{0233} & \textbf{Avg.-8}\\
    \midrule
    \multicolumn{11}{l}{\cellcolor[HTML]{EEEEEE}{\textit{RGB-D Input}}} \\ 
    NICE-SLAM~\cite{zhu2022nice}& 12.0 & 14.0 & 7.9 & 10.9 & 13.4 & \nd 6.2 & 10.7 & \rd 20.9 & 9.0 & 11.8 \\
    Co-SLAM~\cite{Wang_2023_CVPR} & 7.1 & 11.1 & 9.4 & \fst 5.9 & 11.8 & 7.1 & 8.7 & - & - & - \\
    ESLAM~\cite{mahdi2022eslam} & 7.3 & \rd 8.5 & \rd 7.5 & \nd 6.5 & 9.0 & \fst 5.7 & \fst 7.4 & 36.3 & \fst 4.3 & \rd 10.6\\
    MonoGS\cite{matsuki2023gaussian}   & 16.1 & \fst 6.4 & 8.1 & 8.7 & 26.4 & 9.2 & 12.5 & 20.6 & 13.1 & 13.6\\
    \midrule
    \multicolumn{11}{l}{\cellcolor[HTML]{EEEEEE}{\textit{RGB Input}}} \\ 
    MonoGS\cite{matsuki2023gaussian}   & 149.2 & 96.8 & 155.5 & 140.3 & 92.6 & 101.9 & 122.7 & 206.4 & 89.1 & 129.0\\
    GO-SLAM~\cite{zhang2023go}  & \nd 5.9 & \rd 8.3 & 8.1 & 8.4 & \nd 8.3 & \rd 6.9 & \rd 7.7 & \nd 13.3 & \rd 5.3 & \nd 8.1\\
    HI-SLAM\cite{zhang2023hi}   & 6.4 & \nd 7.2 & \fst 6.5 & 8.5 & \fst 7.6 & 8.4 & \fst 7.4 & - & -  & -\\
    Q-SLAM$\textcolor{red}{^*}$\cite{peng2024q}   & \rd 5.8 &  8.5 &  8.4 & 8.7 & \rd 8.8 & - & - & 12.6 & 5.3  & -\\
    GlORIE-SLAM$\textcolor{red}{^*}$~\cite{zhang2024glorie} & \fst 5.5 & 9.1 & \nd 7.0 & \rd 8.2 & \nd 8.3 & 7.5 & \nd 7.6 & \fst 9.4 & \nd 5.1 & \fst 7.5 \\
    \textbf{Ours} & \fst 5.5 & 9.1 & \nd 7.0 & \rd 8.2 & \nd 8.3 & 7.5 & \nd 7.6 & \fst 9.4 & \nd 5.1 & \fst 7.5 \\
    \bottomrule
    \end{tabular}
    \caption{
    \textbf{Tracking Accuracy ATE RMSE [cm] $\downarrow$ on ScanNet~\cite{Dai2017ScanNet}.} Our method equals to GlORIE-SLAM~\cite{zhang2024glorie}, giving the average lowest trajectory error. Results for the RGB-D methods are from \cite{liso2024loopyslam}. Note that all methods with a $\textcolor{red}{^*}$ are concurrent works.
    }
    \label{tab:track_scannet}
\end{table}

\begin{table}[!htb]
    \centering
    \scriptsize
    \setlength{\tabcolsep}{10.5pt}
    \begin{tabularx}{\linewidth}{lccccccc}
    \toprule
    Method  & \texttt{f1/desk} & \texttt{f2/xyz} & \texttt{f3/off}  & \textbf{Avg.-3} & \texttt{f1/desk2} & \texttt{f1/room}& \textbf{Avg.-5}\\
    \midrule
    \multicolumn{8}{l}{\cellcolor[HTML]{EEEEEE}{\textit{RGB-D Input}}} \\ 
    SplaTAM~\cite{keetha2023splatam}& 3.4 & 1.2 & 5.2 & 3.3 & \nd6.5 & 11.1 & \rd5.5\\
    GS-SLAM$\textcolor{red}{^*}$~\cite{yan2023gs} & \rd 1.5 & 1.6 & 1.7 &1.6 & - & - & -\\
    GO-SLAM~\cite{zhang2023go} & \rd 1.5 & \nd0.6 & \nd1.3 &\fst1.1 & - & \nd 4.7 & -\\
    MonoGS~\cite{matsuki2023gaussian} & \nd 1.4 & 1.4 & 1.5 & 1.5 & 5.1 & 6.3 & 3.1 \\
    \midrule
    \multicolumn{8}{l}{\cellcolor[HTML]{EEEEEE}{\textit{RGB Input}}} \\ 
    MonoGS~\cite{matsuki2023gaussian} & 3.8 & 5.2 & 2.9 & 4.0 & 75.7 & 76.6 & 32.8 \\
    Photo-SLAM~\cite{huang2023photo} & \rd1.5 & 1.0 & \nd1.3 & \rd1.3 & - & - & -  \\
    DIM-SLAM~\cite{li2023dense}  & 2.0 & \nd0.6 & 2.3 & 1.6 & - & - & - \\
    GO-SLAM~\cite{zhang2023go} & 1.6 & \nd0.6 &1.5 & \nd1.2 & \nd 2.8 & 5.2 & \nd2.3 \\
    MoD-SLAM$\textcolor{red}{^*}$~\cite{zhou2024modslam} & \rd 1.5 & \rd0.7 & \fst1.1 & \fst1.1 & - & - & - \\
    Q-SLAM$\textcolor{red}{^*}$~\cite{peng2024q} & \fst1.3 & 0.9 & - & - & \fst 2.3 & \rd 4.9 & - \\
    GlORIE-SLAM$\textcolor{red}{^*}$~\cite{zhang2024glorie} & 1.6 & \fst0.2 & \rd1.4 & \fst1.1 & \nd 2.8 & \fst4.2 & \fst2.1 \\
    \textbf{Ours} & 1.6 & \fst0.2 & \rd1.4 & \fst1.1 & \nd 2.8 & \fst4.2 & \fst2.1 \\

    \bottomrule
    \end{tabularx}
    \caption{
    \textbf{Tracking Accuracy ATE RMSE [cm] $\downarrow$ on TUM-RGBD~\cite{Sturm2012ASystems}}. Our method equals to GlORIE-SLAM~\cite{zhang2024glorie}, giving the average lowest trajectory error. Note that all methods with a $\textcolor{red}{^*}$ are concurrent works.
    }
    \label{tab:track_tum_rgbd}
\end{table}

\subsection{Full Evaluations Data}
\begin{table}[!ht]
    \def\dashline{\noalign{\vskip 3pt} \cdashline{2-11}\noalign{\vskip 3pt}}
    \centering
    \scriptsize
    \setlength{\tabcolsep}{6.00pt}
    \resizebox{\columnwidth}{!}
    {
    \begin{tabularx}{\linewidth}{lllccccccccc}
    \toprule
     & & Metric & \texttt{R-0} & \texttt{R-1} & \texttt{R-2} & \texttt{O-0} & \texttt{O-1} & \texttt{O-2} & \texttt{O-3} & \texttt{O-4} & Avg.\\
    \midrule
    \multicolumn{2}{l}{\multirow{5}{*}{\rotatebox{0}{\makecell[l]{Reconstruction}}}}
    & Render Depth L1 $\downarrow$ & 2.90 & 2.16 &2.18 &2.44 &1.97 &2.46 &2.62 & 2.53 & \textbf{2.41}\\
    & &Accuracy $\downarrow$ & 1.99 &1.91 &2.06 &3.96 &2.03 &3.45 &2.15 &1.89 &\textbf{2.43} \\
    & &Completion $\downarrow$ & 3.78 &3.38 &3.34 &2.75 &3.33 &4.36 &3.96 &4.25 & \textbf{3.64}\\
    & &Comp. Rat. $\uparrow$ & 85.47 &86.88 &86.12 &87.32 &85.17 &81.37 &82.25 &82.95 & \textbf{84.69}\\

    \midrule
    \multirow{3}{*}{\rotatebox{0}{\makecell[l]{Rendering}}}
    & \multirow{3}{*}{\rotatebox{0}{\makecell[l]{Keyframes}}}
      &PSNR $\uparrow$&32.25 &34.31 &35.95 &40.81 &40.64 &35.19 &35.03 &37.40 &\textbf{36.45}\\
    & &SSIM $\uparrow$&0.91 & 0.93 &0.95 &0.98 &0.97 &0.96 &0.95 &0.98 & \textbf{0.95}\\
    & &LPIPS $\downarrow$&0.10 &0.09 &0.06 &0.05 &0.05 &0.07 &0.06 &0.04 & \textbf{0.06}\\

\midrule
\renewcommand{\arraystretch}{2}
    \multirow{4}{*}{\rotatebox{0}{\makecell[l]{Tracking}}}
    & \multirow{2}{*}{\makecell[l]{Keyframes\\Trajectory}}
      &\multirow{2}{*}{\makecell[l]{ATE\\ RMSE} $\downarrow$}
      &\multirow{2}{*}{0.29}&\multirow{2}{*}{0.38}&\multirow{2}{*}{0.24}&\multirow{2}{*}{0.27}&\multirow{2}{*}{0.35}&\multirow{2}{*}{0.34}&\multirow{2}{*}{0.42}&\multirow{2}{*}{0.43}&\multirow{2}{*}{\textbf{0.34}}
      \\ \\ \noalign{\vskip 3pt} \cdashline{2-12}\noalign{\vskip 3pt}
    & \multirow{2}{*}{\makecell[l]{Full\\Trajectory}}
      &\multirow{2}{*}{\makecell[l]{ATE\\ RMSE} $\downarrow$}
      &\multirow{2}{*}{0.29}&\multirow{2}{*}{0.33}&\multirow{2}{*}{0.25}&\multirow{2}{*}{0.29}&\multirow{2}{*}{0.35}&\multirow{2}{*}{0.34}&\multirow{2}{*}{0.42}&\multirow{2}{*}{0.43}&\multirow{2}{*}{\textbf{0.34}}
    \\ \\
    \midrule
    \multirow{2}{*}{\makecell[l]{Number of \\ Gaussians}} & \multirow{2}{*}{\makecell[l]{1000x}}
      & 
      &\multirow{2}{*}{116}&\multirow{2}{*}{116}&\multirow{2}{*}{91}&\multirow{2}{*}{76}&\multirow{2}{*}{66}&\multirow{2}{*}{134}&\multirow{2}{*}{114}&\multirow{2}{*}{106}&\multirow{2}{*}{\textbf{102}}
    \\ \noalign{\vskip 6pt}

\bottomrule
\end{tabularx}
    }
\caption{\textbf{Full Evaluation on Replica~\cite{straub2019replica}.} We show the ATE RMSE [cm] evaluation on the keyframes as well as on the full trajectory.} 
\label{tab:replica_full}
\end{table}

\begin{table}[!ht]
    \def\dashline{\noalign{\vskip 3pt} \cdashline{2-11}\noalign{\vskip 3pt}}
    \centering
    \scriptsize
    \setlength{\tabcolsep}{8.00pt}
    \begin{tabularx}{\linewidth}{lllcccccc}
    \toprule
     & & Metric & \texttt{f1/desk} & \texttt{f1/desk2} & \texttt{f1/room} & \texttt{f2/xyz} & \texttt{f3/office} & Avg.\\

    \midrule
    \multirow{3}{*}{\rotatebox{0}{\makecell[l]{Rendering}}}
    & \multirow{3}{*}{\rotatebox{0}{\makecell[l]{Keyframes}}}
      &PSNR $\uparrow$&25.61&23.98&24.06&29.53&26.05&\textbf{25.85}\\
    & &SSIM $\uparrow$&0.84 & 0.81 & 0.80 & 0.90 & 0.84 & \textbf{0.84}\\
    & &LPIPS $\downarrow$&0.18 & 0.23 & 0.24 & 0.08 & 0.20 & \textbf{0.19}\\
    \midrule
    \multirow{2}{*}{\rotatebox{0}{\makecell[l]{Depth \\ Rendering}}}
    & \multirow{2}{*}{\rotatebox{0}{\makecell[l]{Keyframes}}}
    &\multirow{2}{*}{\rotatebox{0}{\makecell[l]{Depth \\ L1$\downarrow$ [cm]}}} &
    \multirow{2}{*}{\rotatebox{0}{\makecell[l]{8.05}}}&
    \multirow{2}{*}{\rotatebox{0}{\makecell[l]{15.70}}}&
    \multirow{2}{*}{\rotatebox{0}{\makecell[l]{15.05}}}&
    \multirow{2}{*}{\rotatebox{0}{\makecell[l]{14.53}}}&
    \multirow{2}{*}{\rotatebox{0}{\makecell[l]{25.59}}}&
    \multirow{2}{*}{\rotatebox{0}{\makecell[l]{\textbf{15.78}}}}\\

     \noalign{\vskip 3pt} \noalign{\vskip 3pt}


\midrule
\renewcommand{\arraystretch}{2}
    \multirow{4}{*}{\rotatebox{0}{\makecell[l]{Tracking}}}
    & \multirow{2}{*}{\makecell[l]{Key Frames\\Trajectory}}
      &\multirow{2}{*}{\makecell[l]{ATE\\ RMSE} $\downarrow$}
      &\multirow{2}{*}{1.92}&\multirow{2}{*}{3.05}&\multirow{2}{*}{4.43}&\multirow{2}{*}{0.23}&\multirow{2}{*}{1.41}&\multirow{2}{*}{\textbf{2.21}}
      \\ \\ \noalign{\vskip 3pt} \cdashline{2-9}\noalign{\vskip 3pt}
    & \multirow{2}{*}{\makecell[l]{Full\\Trajectory}}
      &\multirow{2}{*}{\makecell[l]{ATE\\ RMSE} $\downarrow$}
      &\multirow{2}{*}{1.65}&\multirow{2}{*}{2.79}&\multirow{2}{*}{4.16}&\multirow{2}{*}{0.22}&\multirow{2}{*}{1.44}&\multirow{2}{*}{\textbf{2.05}}
    \\ \\
        \midrule
    \multirow{2}{*}{\makecell[l]{Number of \\ Gaussians}} & \multirow{2}{*}{\makecell[l]{1000x}}
      & 
      &\multirow{2}{*}{88}&\multirow{2}{*}{78}&\multirow{2}{*}{211}&\multirow{2}{*}{173}&\multirow{2}{*}{114}&\multirow{2}{*}{\textbf{133}}
    \\ \noalign{\vskip 6pt}
    
\bottomrule
\end{tabularx}
\caption{\textbf{Full Evaluation on TUM-RGBD~\cite{Sturm2012ASystems}.}} 
\label{tab:tum_full}
    \vspace{-1em}
\end{table}

\begin{table}[!ht]
    \def\dashline{\noalign{\vskip 3pt} \cdashline{2-11}\noalign{\vskip 3pt}}
    \centering
    \scriptsize
    \setlength{\tabcolsep}{6.42pt}
    \begin{tabularx}{\linewidth}{lllccccccccc}
    \toprule
     & & Metric & \texttt{0000} & \texttt{0054} & \texttt{0059} & \texttt{0106} & \texttt{0169} & \texttt{0181} & \texttt{0207} & \texttt{0233} & Avg.\\

    \midrule
    \multirow{3}{*}{\rotatebox{0}{\makecell[l]{Rendering}}}
    & \multirow{3}{*}{\rotatebox{0}{\makecell[l]{Keyframes}}}
    &PSNR$\uparrow$&28.68&30.21&27.69&27.70&31.14&31.15&30.49&27.48&\textbf{29.32}\\
    & &SSIM $\uparrow$&0.83&0.85&0.87&0.86&0.87&0.84&0.84&0.78&\textbf{0.84}\\
    & &LPIPS $\downarrow$&0.19&0.22&0.15&0.18&0.15&0.23&0.19&0.22&\textbf{0.19}\\


\midrule
    \multirow{2}{*}{\rotatebox{0}{\makecell[l]{Depth \\ Rendering}}}
    & \multirow{2}{*}{\rotatebox{0}{\makecell[l]{Keyframes}}}
    &\multirow{2}{*}{\rotatebox{0}{\makecell[l]{Depth \\ L1$\downarrow$ [cm]}}} &
    \multirow{2}{*}{\rotatebox{0}{\makecell[l]{8.24}}}&
    \multirow{2}{*}{\rotatebox{0}{\makecell[l]{18.24}}}&
    \multirow{2}{*}{\rotatebox{0}{\makecell[l]{13.39}}}&
    \multirow{2}{*}{\rotatebox{0}{\makecell[l]{23.5}}}&
    \multirow{2}{*}{\rotatebox{0}{\makecell[l]{11.49}}}&
    \multirow{2}{*}{\rotatebox{0}{\makecell[l]{18.35}}}&
    \multirow{2}{*}{\rotatebox{0}{\makecell[l]{13.78}}}&
    \multirow{2}{*}{\rotatebox{0}{\makecell[l]{10.19}}}&
    \multirow{2}{*}{\rotatebox{0}{\makecell[l]{\textbf{11.37}}}}\\

     \noalign{\vskip 3pt} \noalign{\vskip 3pt}

\midrule
\renewcommand{\arraystretch}{2}
    \multirow{4}{*}{\rotatebox{0}{\makecell[l]{Tracking}}}
    & \multirow{2}{*}{\makecell[l]{Key Frames\\Trajectory}}
      &\multirow{2}{*}{\makecell[l]{ATE\\ RMSE} $\downarrow$}
      &\multirow{2}{*}{5.66}&\multirow{2}{*}{9.17}&\multirow{2}{*}{9.48}&\multirow{2}{*}{7.03}&\multirow{2}{*}{8.72}&\multirow{2}{*}{8.42}&\multirow{2}{*}{7.47}&\multirow{2}{*}{4.97}&\multirow{2}{*}{\textbf{7.61}}
      \\ \\ \noalign{\vskip 3pt} \cdashline{2-12}\noalign{\vskip 3pt}
    & \multirow{2}{*}{\makecell[l]{Full\\Trajectory}}
      &\multirow{2}{*}{\makecell[l]{ATE\\ RMSE} $\downarrow$}
      &\multirow{2}{*}{5.57}&\multirow{2}{*}{9.50}&\multirow{2}{*}{9.11}&\multirow{2}{*}{7.09}&\multirow{2}{*}{8.26}&\multirow{2}{*}{8.39}&\multirow{2}{*}{7.53}&\multirow{2}{*}{5.17}&\multirow{2}{*}{\textbf{7.58}}
    \\ \\

\midrule
    \multirow{2}{*}{\rotatebox{0}{\makecell[l]{Number of\\ Gaussians}}}
    & \multirow{2}{*}{\rotatebox{0}{\makecell[l]{1000x}}}
    &\multirow{2}{*}{\rotatebox{0}{\makecell[l]{}}} &
    \multirow{2}{*}{\rotatebox{0}{\makecell[l]{144}}}&
    \multirow{2}{*}{\rotatebox{0}{\makecell[l]{157}}}&
    \multirow{2}{*}{\rotatebox{0}{\makecell[l]{84}}}&
    \multirow{2}{*}{\rotatebox{0}{\makecell[l]{108}}}&
    \multirow{2}{*}{\rotatebox{0}{\makecell[l]{52}}}&
    \multirow{2}{*}{\rotatebox{0}{\makecell[l]{127}}}&
    \multirow{2}{*}{\rotatebox{0}{\makecell[l]{121}}}&
    \multirow{2}{*}{\rotatebox{0}{\makecell[l]{191}}}&
    \multirow{2}{*}{\rotatebox{0}{\makecell[l]{\textbf{123}}}}\\

     \noalign{\vskip 3pt} \noalign{\vskip 3pt}

\bottomrule
\end{tabularx}
\caption{\textbf{Full Evaluation on ScanNet~\cite{Dai2017ScanNet}.}} 
\label{tab:scannet_full}
    \vspace{-2em}
\end{table}

\noindent
In \cref{tab:replica_full}, \cref{tab:tum_full} and \cref{tab:scannet_full}, we provide the full per scene results on all commonly reported metrics on Replica~\cite{straub2019replica}, TUM-RGBD~\cite{Sturm2012ASystems} and ScanNet~\cite{Dai2017ScanNet}. 

The reconstruction results are only measured on Replica since the other two datasets are real world datasets which lack quality ground truth meshes. 

We show the trajectory accuracy measurement of both keyframes and the full trajectory, which is obtained by first linear interpolation between keyframes and using optical flow to refine. The accuracy of these two trajectories are similar. In the main paper, the data we report is always measured on the full trajectory.

\subsection{Influence of Monocular Depth}

\begin{table}[t]
    \def\dashline{\noalign{\vskip 3pt} \cdashline{2-11}\noalign{\vskip 3pt}}
    \centering
    \scriptsize
    \setlength{\tabcolsep}{7.18pt}
    \resizebox{\columnwidth}{!}
    {
    \begin{tabularx}{\linewidth}{lllccccccccc}
    \toprule
     & & Metric & \texttt{R-0} & \texttt{R-1} & \texttt{R-2} & \texttt{O-0} & \texttt{O-1} & \texttt{O-2} & \texttt{O-3} & \texttt{O-4} & Avg.\\
    \midrule
    \multicolumn{2}{c}{\multirow{5}{*}{\rotatebox{0}{\makecell[l]{Recon-\\struction}}}}
    & Render Depth L1 $\downarrow$ & 2.38 & 1.31 & 1.73 & 1.15 & 1.60 & 1.29 & 5.71 & 1.93 & \textbf{2.14}\\
    & &Accuracy $\downarrow$ & 1.29 & 0.91 & 1.05 & 1.22 & 0.83 & 0.96 & 1.24 & 1.07 & \textbf{1.07}\\
    & &Completion $\downarrow$ & 3.43 & 2.83 & 2.66 & 1.50 & 2.46 & 3.57 & 3.46 & 3.61 & \textbf{2.94}\\
    & &Comp. Rat. $\uparrow$ & 86.61 & 88.69 & 88.70 & 93.44 & 89.09 & 85.20 & 84.60 & 85.32 & \textbf{87.71}\\

    \midrule
    \multicolumn{2}{c}{\multirow{3}{*}{\rotatebox{0}{\makecell[l]{Rendering}}}}
      &PSNR $\uparrow$&35.66 & 37.65 & 38.87 & 43.95 & 43.28 & 37.93 & 37.41 & 39.88& \textbf{39.33} \\
    & &SSIM $\uparrow$&0.96 & 0.96 & 0.97 & 0.99 & 0.98 & 0.96 & 0.96 & 0.98& \textbf{0.97}\\
    & &LPIPS $\downarrow$&0.04 & 0.05 & 0.03 & 0.02 & 0.02 & 0.06 & 0.04 & 0.03& \textbf{0.04}\\

\midrule
\renewcommand{\arraystretch}{2}
    \multirow{2}{*}{\rotatebox{0}{\makecell[l]{Tracking}}}

    &
      &\multirow{2}{*}{\makecell[l]{ATE\\ RMSE} $\downarrow$}
      &\multirow{2}{*}{0.29}&\multirow{2}{*}{0.38}&\multirow{2}{*}{0.24}&\multirow{2}{*}{0.28}&\multirow{2}{*}{0.39}&\multirow{2}{*}{0.35}&\multirow{2}{*}{0.45}&\multirow{2}{*}{0.40}&\multirow{2}{*}{\textbf{0.35}}
    \\ \\

\bottomrule
\end{tabularx}
    }
\caption{\textbf{Full Evaluations on Replica~\cite{straub2019replica} with ground truth depth.} Both reconstruction and rendering results improve significantly with the ground truth depth, suggesting that our method is bounded by the quality of current day monocular depth estimation. Since we do not require any extra training or fine-tuning of the monocular depth estimator, it is easy to plug in a better estimator once available. Tracking performance does not change much.} 
\label{tab:replica_full_rgbd}
    \vspace{-1em}
\end{table}

While we show that the monocular depth improves the geometric estimation capability of our framework, it may still be erroneous. To better understand the accuracy of the monocular depth, we replace it with the ground truth sensor depth instead. This experiment acts as the upper bound of our method if the monocular depth is perfect. The experiments are done on Replica~\cite{straub2019replica} and are shown in \cref{tab:replica_full_rgbd}. Compared with the standard setting with the monocular depth, the ground truth depth setting gives improvements on both reconstruction and rendering quality, which reveals that our method still has potential to achieve better mapping results once better monocular depth is available. Since our method does not require further training or fine-tuning for the monocular depth, it is quite easy to just replace the current off-the-shelf monocular depth estimator with a better one.


\subsection{Impact of Deformation}
During runtime, we deform the 3D Gaussian map to account for adjustments to poses and depth that have already been integrated into the existing map. An alternative to performing the deformation is to solely rely on optimization to resolve the new map. We conduct two experiments to show the benefit of performing the deformation, especially when it comes to rendering accuracy. In \cref{tab:office0_deform}, we vary the number of final refinement iterations and evaluate the rendering depth L1 and PSNR on the Replica \texttt{office 0} scene. We find that utilizing online 3D Gaussian deformations yields better rendering and depth L1 accuracy regardless of the number of iterations. In \cref{tab:scannet_deform} we conduct the same experiment, but over a set of scenes on ScanNet. We find that on average, by enabling the deformation, we achieve higher rendering accuracy and lower depth L1 error. The improvement is, however, more significant when it comes to the rendering accuracy.

\begin{table}[!ht]
    \def\dashline{\noalign{\vskip 3pt} \cdashline{2-11}\noalign{\vskip 3pt}}
    \centering
    \scriptsize
    \setlength{\tabcolsep}{12.00pt}
    \resizebox{\columnwidth}{!}
    {
    \begin{tabularx}{\linewidth}{lllcccc}
    \toprule
    Nbr of Final Iterations $\beta$ & & Metric & \texttt{0K} & \texttt{0.5K} & \texttt{1K} & \texttt{2K}\\
    \midrule
    \multirow{2}{*}{\rotatebox{0}{\makecell[l]{Reconstruction}}} & \multirow{2}{*}{\rotatebox{0}{\makecell[l]{W/O Deform\\W Deform}}}
    & Render Depth L1 $\downarrow$ & 8.84 &	3.49	&2.64	&2.6\\
    & & Render Depth L1 $\downarrow$ & \textbf{6.55}	&\textbf{2.37}&	\textbf{2.34}&	\textbf{2.40}\\

    \midrule
    \multirow{2}{*}{\rotatebox{0}{\makecell[l]{Rendering}}}
    & \multirow{2}{*}{\rotatebox{0}{\makecell[l]{W/O Deform\\W Deform}}}
      &PSNR $\uparrow$& 22.86 &	34.30&	37.66	&37.86\\
      &&PSNR $\uparrow$& \textbf{30.50} &	\textbf{39.87}&	\textbf{40.59}&	\textbf{41.20}\\

\bottomrule
\end{tabularx}
    }
\caption{\textbf{Gaussian Deformation Ablation on Replica~\cite{straub2019replica} \texttt{office 0}}.} 
\label{tab:office0_deform}
\end{table}

\begin{table}[!ht]
    \def\dashline{\noalign{\vskip 3pt} \cdashline{2-11}\noalign{\vskip 3pt}}
    \centering
    \scriptsize
    \setlength{\tabcolsep}{6.42pt}
    \begin{tabularx}{\linewidth}{lllcccccccc}
    \toprule
     & & Metric & \texttt{0000} & \texttt{0054} & \texttt{0059} & \texttt{0106} & \texttt{0169} & \texttt{0181} & \texttt{0207} & Avg.\\

    \midrule
    \multirow{2}{*}{\rotatebox{0}{\makecell[l]{Rendering}}}
    & \multirow{2}{*}{\rotatebox{0}{\makecell[l]{W/O Deform\\W Deform}}}
    &PSNR$\uparrow$&25.15&28.39&\textbf{27.77}&25.25&29.41&30.38&29.30&27.95\\
    &&PSNR$\uparrow$&\textbf{28.68}&\textbf{30.21}&27.69&\textbf{27.70}&\textbf{31.14}&\textbf{31.15}&\textbf{30.49}&\textbf{29.58}\\

\midrule
    \multirow{2}{*}{\rotatebox{0}{\makecell[l]{Depth \\ Rendering}}}
    & \multirow{2}{*}{\rotatebox{0}{\makecell[l]{W/O Deform\\W Deform}}}
    &\multirow{2}{*}{\rotatebox{0}{\makecell[l]{L1$\downarrow$ [cm]}}}
    &
    \multirow{1}{*}{\rotatebox{0}{\makecell[l]{\textbf{7.86}}}}&
    \multirow{1}{*}{\rotatebox{0}{\makecell[l]{22.81}}}&
    \multirow{1}{*}{\rotatebox{0}{\makecell[l]{\textbf{10.51}}}}&
    \multirow{1}{*}{\rotatebox{0}{\makecell[l]{24.19}}}&
    \multirow{1}{*}{\rotatebox{0}{\makecell[l]{11.54}}}&
    \multirow{1}{*}{\rotatebox{0}{\makecell[l]{18.48}}}&
    \multirow{1}{*}{\rotatebox{0}{\makecell[l]{\textbf{13.66}}}}&
    \multirow{1}{*}{\rotatebox{0}{\makecell[l]{15.58}}}\\
    & & &
    \multirow{1}{*}{\rotatebox{0}{\makecell[l]{8.24}}}&
    \multirow{1}{*}{\rotatebox{0}{\makecell[l]{\textbf{18.24}}}}&
    \multirow{1}{*}{\rotatebox{0}{\makecell[l]{13.39}}}&
    \multirow{1}{*}{\rotatebox{0}{\makecell[l]{\textbf{23.5}}}}&
    \multirow{1}{*}{\rotatebox{0}{\makecell[l]{\textbf{11.49}}}}&
    \multirow{1}{*}{\rotatebox{0}{\makecell[l]{\textbf{18.35}}}}&
    \multirow{1}{*}{\rotatebox{0}{\makecell[l]{13.78}}}&
    \multirow{1}{*}{\rotatebox{0}{\makecell[l]{\textbf{15.28}}}}\\

\bottomrule
\end{tabularx}
\caption{\textbf{Gaussian Deformation Ablation on ScanNet~\cite{Dai2017ScanNet}.}} 
\label{tab:scannet_deform}
    \vspace{-2em}
\end{table}

\subsection{Final Refinement Iterations}
After the final global BA step, we perform a final refinement, similar to MonoGS\cite{matsuki2023gaussian}, but include the geometric depth loss as well and do not only refine with a color loss. We ablate the influence on the results by varying the number of iterations of the final refinement in \cref{tab:final_refine_iter}. We find that the rendering accuracy increases monotonically with the number of iterations while the geometric accuracy decreases with more than 2K iterations. We believe this to be a result of fitting to the noisy monocular depth. We choose to use 2K iterations since this provides the best trade-off between rendering and geometric accuracy. 2K iterations takes around 15 seconds on our benchmark hardware which consists of an AMD Ryzen Threadripper Pro 3945WX 12-Cores with an NVIDIA GeForce RTX 3090 Ti with 24 GB of memory.

\begin{table}[!ht]
    \def\dashline{\noalign{\vskip 3pt} \cdashline{2-11}\noalign{\vskip 3pt}}
    \centering
    \scriptsize
    \setlength{\tabcolsep}{12.00pt}
    \resizebox{\columnwidth}{!}
    {
    \begin{tabularx}{\linewidth}{lllcccc}
    \toprule
     Nbr of Final Iterations $\beta$ & & Metric & \texttt{2K} & \texttt{5K} & \texttt{10K} & \texttt{26K}\\
    \midrule
    \multicolumn{2}{l}{\multirow{4}{*}{\rotatebox{0}{\makecell[l]{Reconstruction}}}}
    & Render Depth L1 $\downarrow$ & \textbf{2.36} & 2.45 & 2.51 & 2.59\\
    & &Accuracy $\downarrow$ & \textbf{2.46} & 2.66 & 2.84 & 3.02 \\
    & &Completion $\downarrow$ & 3.60 & \textbf{3.61} & 3.59 & 3.60 \\
    & &Comp. Rat. $\uparrow$ & \textbf{84.87} & 84.71 & 84.80 & 84.77 \\

    \midrule
    \multirow{1}{*}{\rotatebox{0}{\makecell[l]{Rendering}}}
    & \multirow{1}{*}{\rotatebox{0}{\makecell[l]{Keyframes}}}
      &PSNR $\uparrow$& 36.77 & 37.80 & 38.41 & \textbf{38.95}\\

\bottomrule
\end{tabularx}
    }
\caption{\textbf{Final Refinement Iterations Ablation on Replica~\cite{straub2019replica}.} The results are averaged over the 8 scenes.} 
\label{tab:final_refine_iter}
\end{table}

\subsection{Impact of Downsampling Factor}
During mapping, the point cloud formed from unprojecting the depth input is downsampled to avoid adding redundant Gaussians to the scene representation. We investigate the impact of using stronger versus weaker downsampling in \cref{tab:downsampling_ablation} where we also compare to the sensitivity of MonoGS\cite{matsuki2023gaussian} with respect to the same parameter. \Cref{tab:downsampling_ablation} shows that both systems are not very sensitive to the model compression as a result of a larger downsampling factor $\theta$. When both systems use the same number of Gaussians on average ($\theta = 32$ for MonoGS and $\theta = 64$ for our method), we find that our method performs significantly better in terms of depth rerendering and photometric accuracy. For all results in the main paper, we use $\theta = 32$.

\begin{table}[!ht]
    \def\dashline{\noalign{\vskip 3pt} \cdashline{2-11}\noalign{\vskip 3pt}}
    \centering
    \scriptsize
    \setlength{\tabcolsep}{15.00pt}
    \resizebox{\columnwidth}{!}
    {
    \begin{tabularx}{\linewidth}{lllccc}
    \toprule
    Downsampling Factor $\theta$ & & Metric & \texttt{16} & \texttt{32} & \texttt{64} \\
    \midrule
    \multirow{2}{*}{\rotatebox{0}{\makecell[l]{Reconstruction}}}
    &\multirow{2}{*}{\rotatebox{0}{\makecell[l]{Ours \\ MonoGS \cite{matsuki2023gaussian}}}} &\multirow{2}{*}{\rotatebox{0}{\makecell[l]{Render Depth L1 $\downarrow$}}} & 2.38 & 2.40 & 2.46 \\
    & & & 33.43 & 28.47 & 28.09 \\

    \midrule
    \multirow{2}{*}{\rotatebox{0}{\makecell[l]{Rendering}}}
    & \multirow{2}{*}{\rotatebox{0}{\makecell[l]{Ours \\ MonoGS \cite{matsuki2023gaussian}}}}
      &\multirow{2}{*}{\rotatebox{0}{\makecell[l]{PSNR $\uparrow$}}}&36.63 & 36.45 & 36.31  \\
       & &&31.17 & 30.87 & 29.64 \\

    \midrule
    \multirow{2}{*}{\rotatebox{0}{\makecell[l]{Number of \\ Gaussians}}}
    & \multirow{2}{*}{\rotatebox{0}{\makecell[l]{Ours \\ MonoGS \cite{matsuki2023gaussian}}}}
      &\multirow{2}{*}{\rotatebox{0}{\makecell[l]{1000x$\downarrow$}}}&141 &102 & 83  \\
       & &&97 & 83 & 73 \\

\bottomrule
\end{tabularx}
    }
\caption{\textbf{Downsampling Factor $\theta$ Ablation on Replica~\cite{straub2019replica}.} The results are averaged over the 8 scenes.} 
\label{tab:downsampling_ablation}
\end{table}

\subsection{Runtime Evaluation}
To be consistent with the keyframe selection hyperparameters of MonoGS~\cite{matsuki2023gaussian}, we report on the same parameters as MonoGS uses by default. In practice, this means that few keyframes from the tracking system (determined via mean optical flow thresholding) are actually filtered out and not mapped. In \cref{tab:speed_up_runtime}, we show that by altering the hyperparamters, we can speed up the system during runtime, while still rendering and reconstructing the scene well. Note that we evaluate the rendering performance on the same set of views for all runs. We benchmark the runtime on an AMD Ryzen Threadripper Pro 3945WX 12-Cores with an NVIDIA GeForce RTX 3090 Ti with 24 GB of memory. We note that we currently do not leverage multiprocessing to the amount possible in practice \ie currently we first do tracking and then mapping \ie there is no simultaneous tracking and mapping. This is, however, straightforward to include, which should further speed up the runtime.

\begin{table}[!ht]
    \def\dashline{\noalign{\vskip 3pt} \cdashline{2-11}\noalign{\vskip 3pt}}
    \centering
    \scriptsize
    \setlength{\tabcolsep}{5.00pt}
    \resizebox{\columnwidth}{!}
    {
    \begin{tabularx}{\linewidth}{llccccccc}
    \toprule
     \(k_{\text{fcov}}\), \(k_{\text{fm}}\) &  & 0.95, 0.04 & 0.90, 0.08 & 0.85, 0.08 & 0.80, 0.12 & 0.70, 0.16 &  0.60, 0.20 & 0.50, 0.30\\
    \midrule
    \multirow{4}{*}{\rotatebox{0}{\makecell[l]{Reconstruction}}}
    & Render Depth L1 $\downarrow$ & \textbf{2.90} & 2.94 & 2.97 & 3.08 & 3.37 & 3.53 & 4.78\\
    & Accuracy $\downarrow$ & 1.99 & \textbf{1.94} & 2.06 & 2.04 & 2.54 & 3.20 & 6.20 \\
    & Completion $\downarrow$ & 3.78 & \textbf{3.76} & 3.79 & 3.77 & 3.86 & 3.93 & 5.23 \\
    & Comp. Rat. $\uparrow$ & 85.47 & \textbf{85.58} & 85.39 & 85.53 & 85.03 & 84.33 & 80.38 \\

    \midrule
    \multirow{1}{*}{\rotatebox{0}{\makecell[l]{Rendering}}}

      &PSNR $\uparrow$& \textbf{32.25} & 31.65 & 31.31 & 30.59 & 30.12 & 29.25 & 27.59\\

    \midrule
    \multirow{1}{*}{\rotatebox{0}{\makecell[l]{Runtime}}}
   
      &FPS $\uparrow$& 1.24 & 1.45 & 1.62 & 2.02 & 2.50 & 3.03 & \textbf{3.67} \\

\bottomrule
\end{tabularx}
    }
\caption{\textbf{Keyframe Hyperparameter Search on Replica~\cite{straub2019replica} \texttt{room 0}.} By changing the keyframe selection hyperparameters, we can speed up our runtime without impacting reconstruction and rendering too much. We evaluate the rendering performance on the same set of frames for all runs. In comparison, with the default \(k_{\text{fcov}} = 0.95\), \(k_{\text{fm}} = 0.04 \), MonoGS \cite{matsuki2023gaussian} yields PSNR: 26.12 and render depth L1: 17.38 cm.} 
\label{tab:speed_up_runtime}
\end{table}

\subsection{Additional Qualitative Reconstructions}
\begin{figure}[ht]
\vspace{0em}
\centering
{
\setlength{\tabcolsep}{1pt}
\renewcommand{\arraystretch}{1}
\newcommand{\sz}{0.23}
\newcommand{\subsz}{0.2}
\begin{tabular}{ccccc}
\raisebox{0.35cm}{\rotatebox{90}{\textbf{\makecell{\texttt{Office 4}}}}}&
\includegraphics[width=\sz\linewidth]{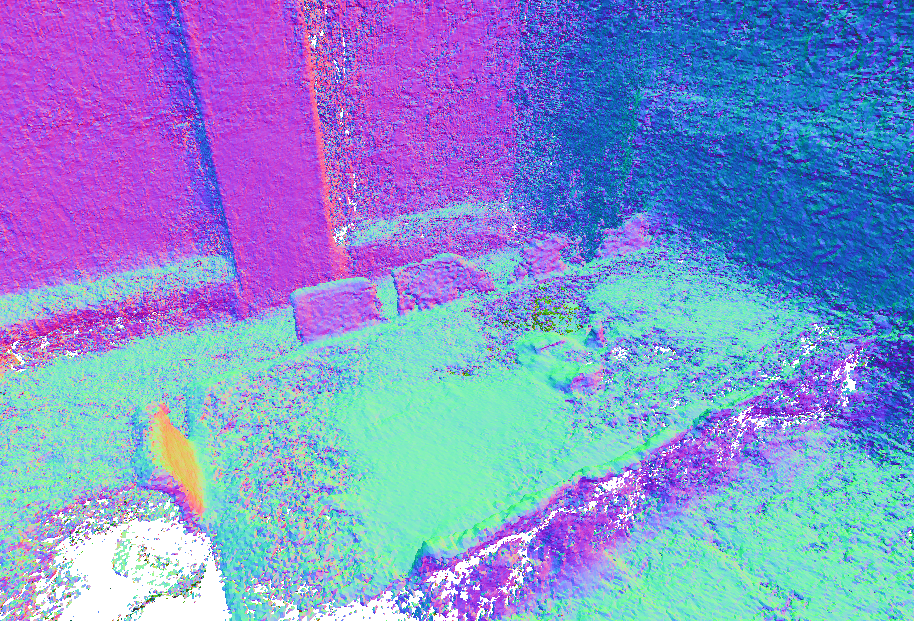} &
\includegraphics[width=\sz\linewidth]{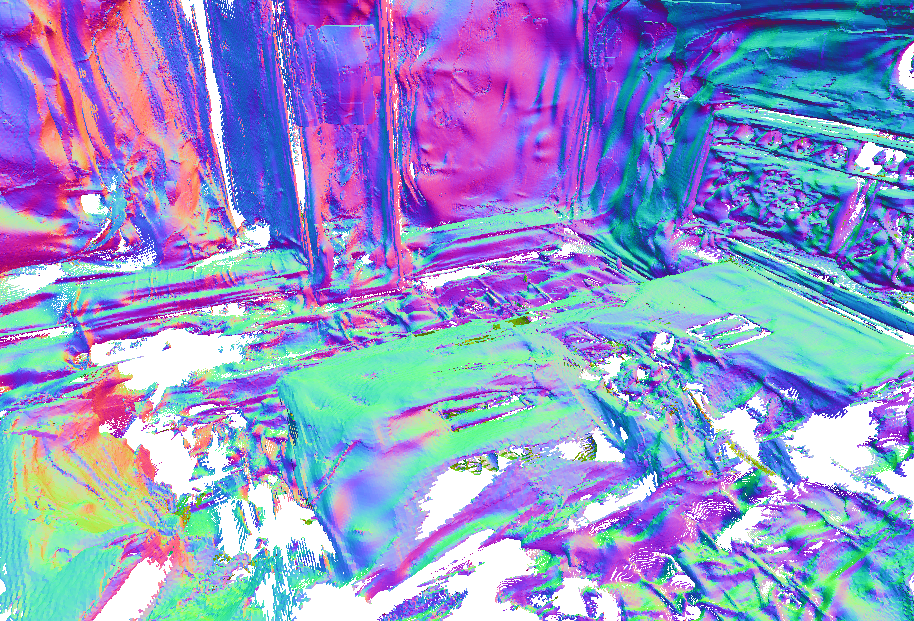} &
\includegraphics[width=\sz\linewidth]{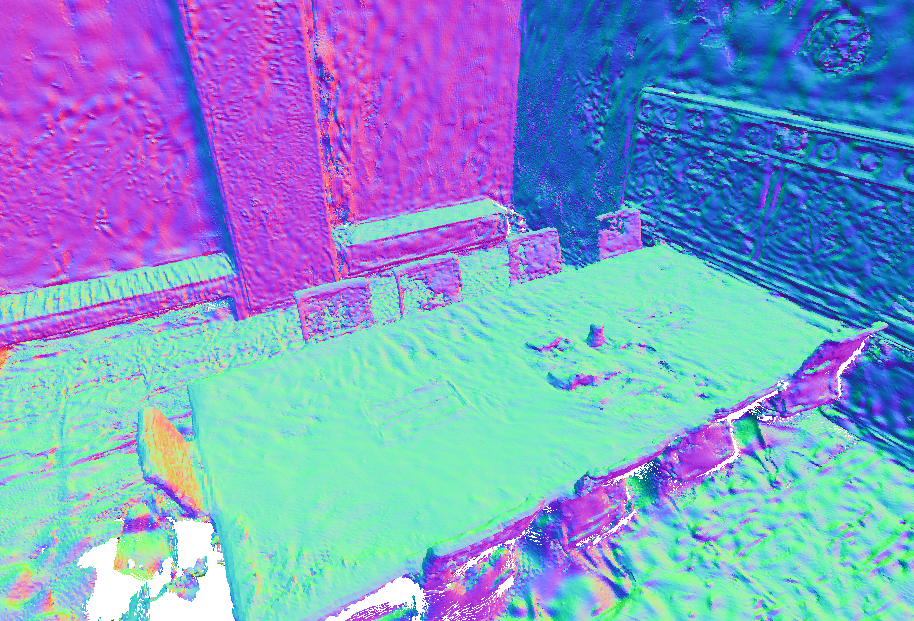} &
\includegraphics[width=\sz\linewidth]{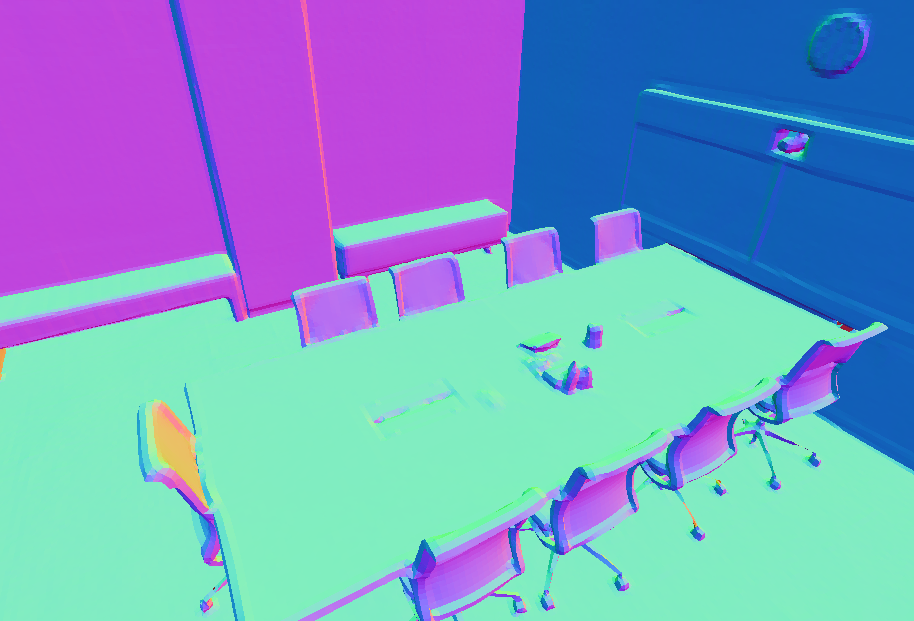} 
\\

\Large
& GlORIE-SLAM$\textcolor{red}{^*}$~\cite{zhang2024glorie}  & MonoGS~\cite{matsuki2023gaussian} & \ours (Ours)  & Ground Truth \\
\end{tabular}
}
\caption{\textbf{Reconstruction Results on Replica~\cite{straub2019replica}.} Our method improves upon the geometric accuracy compared to existing works, when observing the normal shaded meshes. In particular, GlORIE-SLAM suffers from floating point artifacts. MonoGS suffers badly from a lack of proxy depth, despite multiview optimization.
\label{fig:reconstruction_replica_supp}}
\vspace{0em}
\end{figure}

In \cref{fig:reconstruction_replica_supp} we show additional qualitative results from the Replica dataset on normal shaded meshes.

\clearpage

{\small
\bibliographystyle{splncs04}
\bibliography{main}
}

\end{document}